\documentclass{article} % For LaTeX2e
\usepackage{iclr2024_conference,times}

\usepackage{amsmath,amsfonts,bm}

\def\eqref#1{equation~\ref{#1}}
\def\1{\bm{1}}

\DeclareMathAlphabet{\mathsfit}{\encodingdefault}{\sfdefault}{m}{sl}
\SetMathAlphabet{\mathsfit}{bold}{\encodingdefault}{\sfdefault}{bx}{n}

\newcommand{\E}{\mathbb{E}}

\usepackage{hyperref}
\usepackage{url}

\usepackage[utf8]{inputenc} % allow utf-8 input
\usepackage[T1]{fontenc}    % use 8-bit T1 fonts
\usepackage{hyperref}       % hyperlinks
\usepackage{url}            % simple URL typesetting
\usepackage{booktabs}       % professional-quality tables
\usepackage{amsfonts}       % blackboard math symbols
\usepackage{nicefrac}       % compact symbols for 1/2, etc.
\usepackage{microtype}      % microtypography
\usepackage{xcolor}         % colors

\usepackage{arydshln} % for dash line in table
\usepackage{graphicx}
\usepackage{subfigure}
\usepackage{capt-of}
\usepackage{array}
\usepackage{wrapfig}
\usepackage{makecell}
\usepackage{amssymb}
\usepackage{amsmath}
\usepackage{multirow}
\usepackage{ccaption}
\usepackage{caption}
\usepackage{pifont} % checkmark and x mark
\usepackage{romannum} % for roman number
\usepackage{algorithm} % for algorithm
\usepackage{algpseudocode} % for algorithm
\usepackage{dsfont}
\usepackage{arydshln} % for dash line in table
\usepackage{xcolor}
\title{Causal Unsupervised Semantic Segmentation}

\iclrfinalcopy
\author{%
  Junho Kim\thanks{Equal contribution. $\dagger$ Corresponding author.},~~Byung-Kwan Lee\footnote[1]{},~~Yong Man Ro\footnote[2]{}\\
  School of Electrical Engineering \\
  Korea Advanced Institute of Science and Technology (KAIST) \\
  \texttt{\{arkimjh,~leebk,~ymro\}@kaist.ac.kr} \\
}

\begin{document}

\pagenumbering{arabic} % page is arabic number, it is not NeurIPS22 pacakage

\maketitle

\begin{abstract}
Unsupervised semantic segmentation aims to achieve high-quality semantic grouping without human-labeled annotations. With the advent of self-supervised pre-training, various frameworks utilize the pre-trained features to train prediction heads for unsupervised dense prediction. However, a significant challenge in this unsupervised setup is determining the appropriate level of clustering required for segmenting concepts. To address it, we propose a novel framework, CAusal Unsupervised Semantic sEgmentation (CAUSE), which leverages insights from causal inference. Specifically, we bridge intervention-oriented approach (\textit{i.e.,} frontdoor adjustment) to define suitable two-step tasks for unsupervised prediction. The first step involves constructing a concept clusterbook as a mediator, which represents possible concept prototypes at different levels of granularity in a discretized form. Then, the mediator establishes an explicit link to the subsequent concept-wise self-supervised learning for pixel-level grouping. Through extensive experiments and analyses on various datasets, we corroborate the effectiveness of CAUSE and achieve state-of-the-art performance in unsupervised semantic segmentation.
\end{abstract}

% Introduction Start
\section{Introduction}

\label{sec:1} 
Semantic segmentation is one of the essential computer vision tasks that has continuously advanced in the last decade with the growth of Deep Neural Networks (DNNs)~\citep{resnet, dosovitskiy2020image, carion2020end} and large-scale annotated datasets~\citep{everingham2010pascal, cordts2016cityscapes, caesar2018coco}. However, obtaining such pixel-level annotations for dense prediction requires an enormous amount of human resources and is more time-consuming compared to other image analysis tasks. Alternatively, weakly-supervised semantic segmentation approaches have been proposed to relieve the costs by using of facile forms of supervision such as class labels~\citep{wang2020self, zhang2020causal}, scribbles~\citep{lin2016scribblesup}, bounding boxes~\citep{dai2015boxsup, khoreva2017simple}, and image-level tags~\citep{xu2015learning, tang2018regularized}.

While relatively few works have been dedicated to explore unsupervised semantic segmentation (USS), several methods have presented the way of segmenting feature representations without any annotated labels by exploiting visual consistency maximization~\citep{ji2019invariant, hwang2019segsort}, multi-view equivalence~\citep{cho2021picie}, or saliency priors~\citep{van2021unsupervised, ke2022unsupervised}. In parallel with segmentation researches, recent self-supervised learning frameworks~\citep{caron2021emerging, bao2022beit} using Vision Transformer have observed that their representations exhibit semantic consistency at the pixel-level scale for object targets. Based on such intriguing properties of self-supervised training, recent USS methods~\citep{hamilton2022unsupervised, ziegler2022self, yin2022transfgu, zadaianchuk2023unsupervised, liacseg, seong2023leveraging} have employed the pre-trained features as a powerful source of prior knowledge and introduced contrastive learning frameworks by maximizing feature correspondence for the unsupervised segmentation task.

%%%%%%%%%%%%%%%%%%%%%%%%%%%%%%%%%%%%%%%%%%%%%%%%%%%%%%%%%%%%%%%%%%%%%%%%%%%%%%%%%%%%
\begin{figure}[t]
\vspace{-13mm}
\centering
\includegraphics[width=0.99\textwidth]{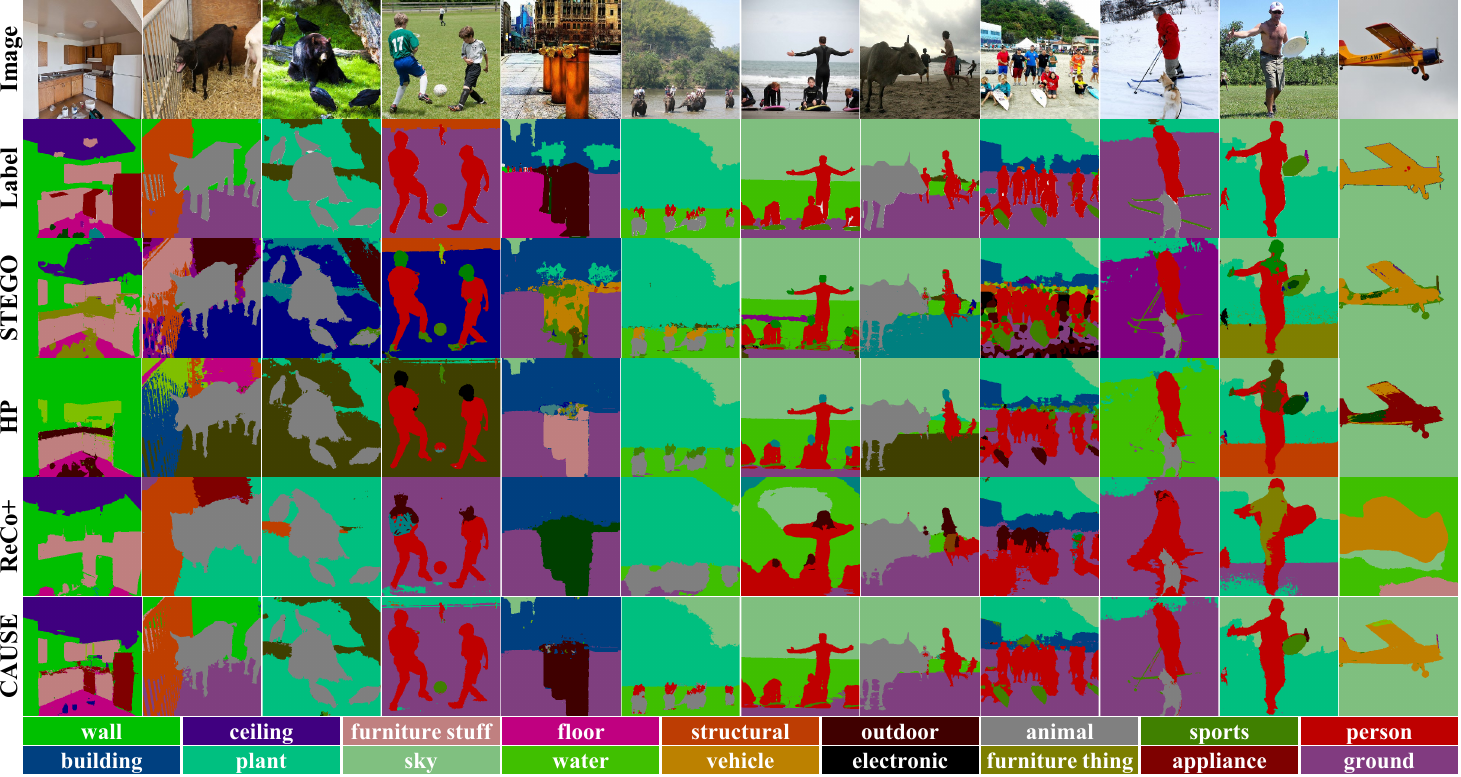}
\vspace*{-2.5mm}
\caption{Visual comparison of USS for COCO-stuff~\citep{caesar2018coco}. Note that, in contrast to true labels, baseline frameworks~\citep{hamilton2022unsupervised, seong2023leveraging, shin2022reco} fail to achieve targeted level of granularity, while CAUSE successfully clusters \textit{person, sports, vehicle}, etc.}
\label{fig:1}
\vspace{-5mm}
\end{figure}
%%%%%%%%%%%%%%%%%%%%%%%%%%%%%%%%%%%%%%%%%%%%%%%%%%%%%%%%%%%%%%%%%%%%%%%%%%%%%%%%%%%%

\begin{wrapfigure}{h}{0.34\textwidth}
\vskip -2.9ex
\centering
\includegraphics[width=0.9\linewidth]{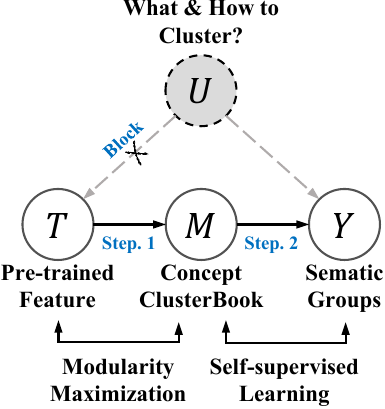}
% \vskip 0.5ex
\vspace{-3.2mm}
\caption{Causal diagram of CAUSE. We split USS into two steps to identify relation between pre-trained features $T$ and semantic groups $Y$ using \textit{clusterbook} $M$.}
\vspace{-2mm}
\label{fig:causal}
\end{wrapfigure}

In this paper, we begin with a fundamental question for the unsupervised semantic segmentation: \textit{How can we define what to cluster and how to do so under an unsupervised setting?}, which has been overlooked in previous works. A major challenge for USS lies in the fact that unsupervised segmentation is more akin to clustering rather than semantics with respect to pixel representation. Therefore, even with the support of self-supervised representation, the lack of awareness regarding what and how to cluster for each pixel representation makes USS a challenging task, especially when aiming for the desired level of granularity. For example, elements such as \textit{head, torso, hand, leg, etc.,} should ideally be grouped together under the broader-level category \textit{person}, a task that previous methods~\citep{hamilton2022unsupervised, seong2023leveraging} have had difficulty accomplishing, as in Fig.~\ref{fig:1}. To address these difficulties, we, for the first time, treat USS procedure within the context of causality and propose suitable two-step tasks for the unsupervised learning. As shown in Fig.~\ref{fig:causal}, we first schematize a causal diagram for a simplified understanding of causal relations for the given variables and the corresponding unsupervised tasks for each step. Note that our main goal is to group semantic concepts $Y$ that meet the targeted level of granularity, utilizing feature representation $T$ from pre-trained self-supervised methods such as DINO~\citep{caron2021emerging}.

%%%%%%%%%%%%%%%%%%%%%%%%%%%%%%%%%%%%%%%%%%%%%%%%%%%%%%%%%%%%%%%%%%%%%%%%%%%%%%
%%%%%%%%%%%%%%%%%%%%%%%%%%%%%%%%%%%%%%%%%%%%%%%%%%%%%%%%%%%%%%%%%%%%%%%%%%%%%%

Specifically, the unsupervised segmentation ($T \rightarrow Y$) is a procedure for deriving semantically clustered groups $Y$ distilled from pre-trained features $T$. However, the indeterminate $U$ of unsupervised prediction (\textit{i.e.,} what and how to cluster) can lead confounding effects during pixel-level clustering without supervision. Such effects can be considered as a backdoor path ($T \leftarrow U \rightarrow Y$) that hinders the targeted level of segmentation. Accordingly, our primary insight stems from constructing a subdivided concept representation $M$, with discretized indices, which serves as an explicit link between $T$ and $Y$ in alternative forms of supervision. Intuitively, the construction of subdivided concept \textit{clusterbook} $M$ implies the creation of as many inherent concept prototypes as possible in advance, spanning various levels of granularity. Subsequently, for the given pre-trained features, we train a segmentation head that can effectively consolidate the concept prototypes into the targeted broader-level categories using the constructed clusterbook. This strategy involves utilizing the discretized indices within $M$ to identify positive and negative features for the given anchor, enabling concept-wise self-supervised learning.

Beyond the intuitive causal procedure of USS, building a mediator $M$ can be viewed as a blocking procedure of the backdoor paths induced from $U$ by assigning possible concepts in discretized states such as in~\citet{van2017neural, esser2021taming}. That is, it satisfies a condition for frontdoor adjustment~\citep{pearl1993bayesian}, which is a powerful causal estimator that can establish only causal association\footnote{In Step 1, $Y$ is a collider variable in the path of $T {\rightarrow} Y$ through $U$, and it blocks backdoor path. Therefore, causal association only flows into $M$ from $T$. Then, in Step 2, $T$ blocks $M {\leftarrow} T {\leftarrow} U {\rightarrow} Y$. By combining two steps, we can distill the pre-trained representation using only causal association path and reflect it on semantic groups, which is our ultimate goal for unsupervised semantic segmentation. Please see preliminary in Section~\ref{sec:dgp}.} ($T \rightarrow M \rightarrow Y$). We name our novel framework as CAusal Unsupervised Semantic sEgmentation (CAUSE), which integrates the causal approach into the field of USS. As illustrated in Fig.~\ref{fig:causal}, in brief, we divide the unsupervised dense prediction into two step tasks: (1) discrete subdivided representation learning with \textit{Modularity} theory~\citep{newman2006modularity} and (2) conducting do-calculus~\citep{pearl1995causal} with self-supervised learning~\citep{oord2018representation} in the absence of annotations. By combining the above tasks, we can bridge causal inference into the unsupervised segmentation and obtain semantically clustered groups with the support of pre-trained feature representation.

Our main contributions can be concluded as: (\lowercase\expandafter{\romannumeral1}) We approach unsupervised semantic segmentation task with an intervention-oriented approach (\textit{i.e.,} causal inference) and propose a novel unsupervised dense prediction framework called CAusal Unsupervised Semantic sEgmentation (CAUSE), (\lowercase\expandafter{\romannumeral2}) 
To address the ambiguity in unsupervised segmentation, we integrate frontdoor adjustment into USS and introduce two-step tasks: deploying a discretized concept clustering and concept-wise self-supervised learning, and (\lowercase\expandafter{\romannumeral3}) Through extensive experiments, we corroborate the effectiveness of CAUSE on various datasets and achieve state-of-the-art results in unsupervised semantic segmentation.

\section{Related Work}
\label{sec:2}
As an early work for USS, \citet{ji2019invariant} have proposed IIC to maximize mutual information of feature representations from augmented views. After that, several methods have further improved the segmentation quality by incorporating inductive bias in the form of cross-image correspondences~\citep{hwang2019segsort,cho2021picie,wen2022selfsupervised} or saliency information in an end-to-end manner~\citep{van2021unsupervised, ke2022unsupervised}. Recently, with the discovery of semantic consistency for pre-trained self-supervised frameworks~\citep{caron2021emerging}, \citet{hamilton2022unsupervised} have leveraged the pre-trained features for the unsupervised segmentation. Subsequently, various works~\citep{wen2022selfsupervised, yin2022transfgu, ziegler2022self} have utilized the self-supervised representation as a form of pseudo segmentation labels~\citep{zadaianchuk2023unsupervised, liacseg} or a pre-encoded representation to further incorporate additional prior knowledge~\citep{van2021unsupervised, zadaianchuk2023unsupervised} into the segmentation frameworks. Our work aligns with previous studies~\citep{hamilton2022unsupervised, seong2023leveraging} in the aspect of refining segmentation features using pre-trained representations without external information. However, we highlight that the lack of a well-defined clustering target in the unsupervised setup leads to suboptimal segmentation quality. Accordingly, we interpret USS within the context of causality, bridging the construction of discretized representation with pixel-level self-supervised learning (see extended explanations in Appendix \ref{appendix:A}.)

% Method Start
\section{Causal Unsupervised Semantic Segmentation}
\label{sec:3}

%%%%%%%%%%%%%%%%%%%%%%%%%%%%%%%%%%%%%%%%%%%%%%%%%%%%%%%%%%%%%%%%%%%%%%%%%%%%%%%%%%%%
\begin{figure*}[t]
\vspace{-13mm}
\centering
\includegraphics[width=0.99\textwidth]{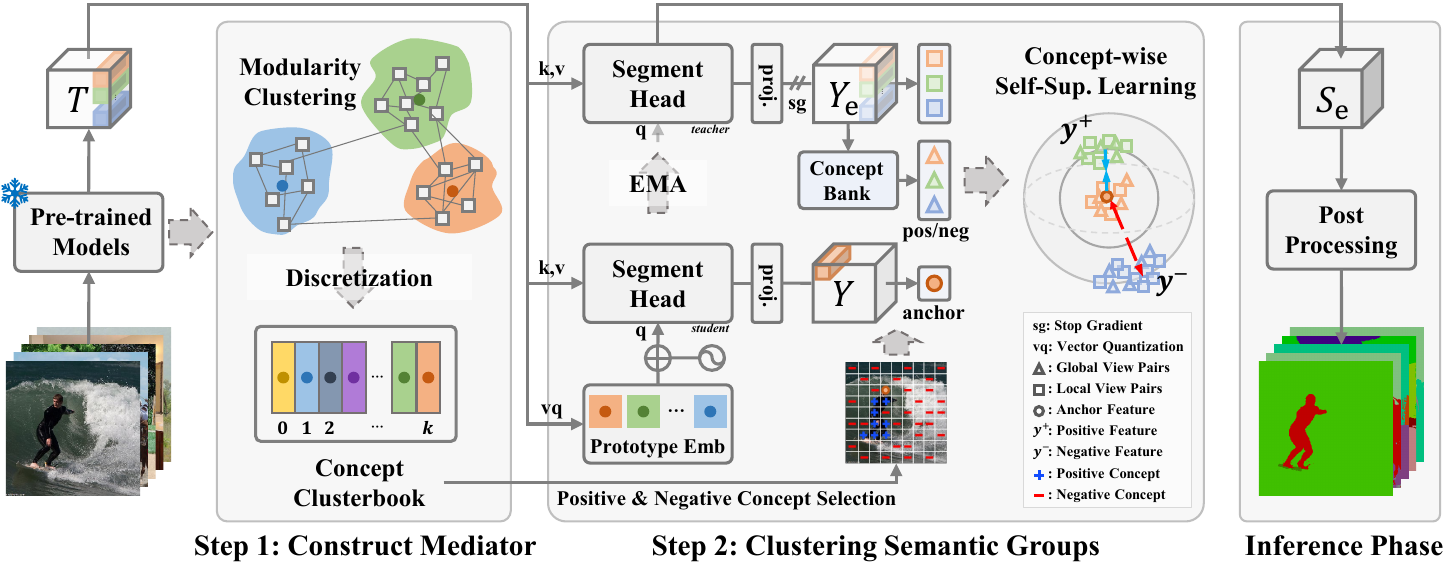}
\vspace*{-3mm}
\caption{The overall architecture of CAUSE comprises (\lowercase\expandafter{\romannumeral1}): constructing discretized concept clusterbook as a mediator and (\lowercase\expandafter{\romannumeral2}): clustering semantic groups using concept-wise self-supervised learning.}
\vspace{-5mm}
\label{fig:3}
\end{figure*}
%%%%%%%%%%%%%%%%%%%%%%%%%%%%%%%%%%%%%%%%%%%%%%%%%%%%%%%%%%%%%%%%%%%%%%%%%%%%%%%%%%%%

\subsection{Data Generating Process for Unsupervised Semantic Segmentation}
\label{sec:dgp}
\paragraph{Preliminary.} It is important to define Data Generating Process (DGP) early in the process for causal inference. DGP outlines the causal relationships between treatment $T$ and outcome of our interest $Y$, and the interrupting factors, so-called confounder $U$. For example, if we want to identify the causal relationship between \textit{smoking} (\textit{i.e.,} treatment) and \textit{lung cancer} (\textit{i.e.,} outcome of our interest), \textit{genotype} can be deduced as one of potential confounders that provoke confounding effects between \textit{smoking} $T$ and \textit{lung cancer} $Y$.
Once we define the confounder $U$, and if it is observable, backdoor adjustment~\citep{pearl1993bayesian} is an appropriate solution to estimate the causal influence between $T$ and $Y$ by controlling $U$. However, not only in the above example but also in many real-world scenarios, including high-dimensional complex DNNs, confounder is often unobservable and either uncontrollable. In this condition, controlling $U$ may not be a feasible option, and it prevents us from precisely establishing the causal relationship between $T$ and $Y$.

Fortunately, \citet{pearl2009causality} introduces frontdoor adjustment allowing us to elucidate the causal association even in the presence of unobservable confounder $U$. Here, the key successful points for frontdoor adjustment are two factors, as shown in Fig.~\ref{fig:causal}: (a) assigning a mediator $M$ bridging treatment $T$ into outcome of our interest $Y$ while being independent with confounder $U$ and (b) averaging all possible treatments between the mediator and outcome. When revisiting the above example, we can instantiate a mediator $M$ as accumulation of \textit{tar} in lungs, which only affects \textit{lung cancer} $Y$ from \textit{smoking} $T$. We then average the probable effect between \textit{tar} $M$ and \textit{lung cancer} $Y$ across all of the participants' population $T'$ on \textit{smoking}. The following formulation represents frontdoor adjustment:
\begin{equation}
\label{eqn:frontdoor}
    p(Y\mid \text{do}(T) )=\underbrace{\sum_{m\in M}p(m \mid T)}_{\text{Step 1}}\underbrace{\sum_{t'\in T'}{p(Y\mid t', m)}p(t')}_{\text{Step 2}},
\end{equation}
where $\text{do}(\cdot)$ operator describes do-calculus~\citep{pearl1995causal}, which indicates intervention on treatment $T$ to block unassociated backdoor path induced from $U$ between the treatments and outcome of interest.

\paragraph{Causal Perspective on USS.} Bridging the causal view into unsupervised semantic segmentation, our objective is clustering semantic groups $Y$ with a support of pre-trained self-supervised features $T$. Here, in unsupervised setups, we define $U$ as indetermination during clustering (\textit{i.e.,} a lack of awareness about what and how to cluster), which cannot be observed within the unsupervised context. Therefore, in Step 1 of Eq.~(\ref{eqn:frontdoor}), we first need to build a mediator directly relying on $T$ while being independent with the unobserved confounder $U$. To do so, we construct concept clusterbook as $M$, which is set of concept prototypes that encompass potential concept candidates spanning different levels of granularity only through $T$. The underlying assumption for the construction of $M$ is based on the object alignment property observed in recent self-supervised methods~\citep{caron2021emerging, oquab2023dinov2}, a characteristic exploited by~\citet{hamilton2022unsupervised, seong2023leveraging}. Next, in Step 2 of Eq.~(\ref{eqn:frontdoor}), we need to determine whether to consolidate or separate the concept prototypes into the targeted semantic-level groups $Y$. We utilize the discretized indices from $M$ for discriminate positive and negative features for the given anchor and conduct concept-wise self-supervised learning. The following is an approximation of Eq.~(\ref{eqn:frontdoor}) for the unsupervised dense prediction:
\begin{equation}
\label{eqn:frontdoor2}
    \mathop{\mathbb{E}}_{t\in T}\left[p(Y\mid \text{do}(t) )\right]=\mathop{\mathbb{E}}_{t\in T}\left[\sum_{m\in M}p(m \mid t)\sum_{t'\in T'}{p(Y\mid t', m)}p(t')\right],
\end{equation}
where, $T'$ indicates a population of all feature points, but notably in a pixel-level manner suitable for dense prediction. In summary, our focus is enhancing $p(Y|\text{do}(t))$ for feature points $t$ by assigning appropriate unsupervised two tasks (\lowercase\expandafter{\romannumeral1}) $p(m|t)$: construction of concept clusterbook and (\lowercase\expandafter{\romannumeral2}) $p(Y|t', m)$: concept-wise self-supervised learning, all of which can be bridged to frontdoor adjustment.

\begin{figure}[t!]
\vspace{-12mm}
\begin{algorithm}[H]
\caption{(STEP 1) Maximizing Modularity for Constructing Concept Clusterbook $M$}
\begin{algorithmic}[1]
\Require Image Samples $X\sim \text{Data}$, Pre-trained Model $f$, Concept Fractions $M\in\mathbb{R}^{k\times c}$
\State \text{Initialize} $M$
\For {$X \sim\text{Data}$}
\State $T\in\mathbb{R}^{hw\times c} \gets f(X)$ \Comment{Pre-trained Model Representation}
\State $\mathcal{A}\gets\max(0, \text{cos}(T, T))\in\mathbb{R}^{hw\times hw}$ \Comment{Affinity matrix}
\State $d, e\gets\mathcal{A}$ \Comment{Degree Matrix and Number of Total edges}
\State $\mathcal{C}\gets\max(0, \text{cos}(T, M))\in\mathbb{R}^{hw\times k}$ \Comment{Cluster Assignment Matrix}
\State $\mathcal{H}\gets \frac{1}{2e}\text{Tr}\left(\tanh\left(\frac{\mathcal{C}\mathcal{C}^T}{\tau}\right)\left[\mathcal{A}-\frac{dd^T}{2e}\right]\right)$ \Comment{Maximizing Modularity ($\tau=0.1$)}
\State $M \gets \text{Increase}(\mathcal{H})$ \Comment{Updating Concept ClusterBook (lr: $0.001$)}
\EndFor
\end{algorithmic}
\label{alg:mediator}
\end{algorithm}
\vspace{-7mm}
\end{figure}

\subsection{Constructing Concept Clusterbook for Mediator}
\label{sec:3.2}
\paragraph{Concept Prototypes.} We initially define a mediator $M$ and maintain it as a link between the pre-trained features $T$ and the semantic groups $Y$. This mediator necessitates an explicit representation that transforms the continuous representation found in pre-trained self-supervised frameworks, into a discretized form. One of possible approaches is reconstruction-based vector-quantization~\citep{van2017neural, esser2021taming} that is well-suited for generative modeling. However, for dense prediction, we require more sophisticated pixel-level clustering methods that consider pixel locality and connectivity. Importantly, they should be capable of constructing such representations in discretized forms for alternative role of supervisions. Accordingly, we exploit a clustering method that maximizes modularity~\citep{newman2006modularity}, which is one of the most effective approaches for considering relations among vertices. The following formulation represents maximizing a measure of modularity $\mathcal{H}$ to acquire the discretized concept fractions from pre-trained features $T$:
% be capable of constructing representations in discretized forms
\begin{equation}
    \max_{M}\mathcal{H} = \frac{1}{2e}\text{Tr}\left(\mathcal{C}(T, M)^{\text{T}}\left[\mathcal{A}(T)-\frac{dd^{\text{T}}}{2e}\right]\mathcal{C}(T, M)\right)\in\mathbb{R},
\label{eqn:modularity}
\end{equation}
where $\mathcal{C}(T, M)\in\mathbb{R}^{hw\times k}$ denotes cluster assignment matrix such that $\max(0, \text{cos}(T, M))$ between all $hw$ patch feature points in pre-trained features $T\in\mathbb{R}^{hw \times c}$ and all $k$ concept prototypes in $M\in\mathbb{R}^{k\times c}$. The cluster assignment matrix implies how each patch feature point is close to concept prototypes. In addition, $\mathcal{A}(T)\in\mathbb{R}^{hw\times hw}$ indicates the affinity matrix of $T=\{t\in\mathbb{R}^{c}\}^{hw}$ such that $\mathcal{A}_{ij}=\max(0, \text{cos}(t_i, t_j))$ between the two patch feature points $t_i, t_j$ in $T$, which represents the intensity of connections among vertices. Note that, degree vector $d\in\mathbb{R}^{hw}$ describes the number of the connected edges in its affinity $\mathcal{A}$, and $e\in\mathbb{R}$ denotes the total number of the edges. 

By jointly considering cluster assignments $\mathcal{C}(T,M)$ and affinity matrix $\mathcal{A}(T)$ at once, in brief, maximizing modularity $\mathcal{H}$ constructs the discretized concept clusterbook $M$ taking into account the patch-wise locality and connectivity in pre-trained representation $T$. In practical, directly calculating Eq.~(\ref{eqn:modularity}) can lead to much small value of $\mathcal{H}$ due to multiplying tiny elements of $\mathcal{C}$ twice. Thus, we use trace property and hyperbolic tangent with temperature term $\tau$ to scale up $\mathcal{C}$ (see Appendix \ref{appendix:B}). Algorithm~\ref{alg:mediator} provides more details on achieving maximizing modularity to generate concept clusterbook $M$, where we train only one epoch with Adam~\citep{kingma2014adam} optimizer.

\subsection{Enhancing Likelihood for Semantic Groups}
\label{sec:3.3}

\paragraph{Concept-Matched Segmentation Head.} As part of Step 2, to embed segmentation features $Y$ that match with concept prototypes from pre-trained features $T$, we train a task-specific prediction head $S$. As in Fig.~\ref{fig:3}, the pre-trained model remains frozen, and their features $T=\{t\in\mathbb{R}^{c}\}^{hw}$ are fed into the segmentation head $S$ that performs cross-attention with querying prototype embedding $Q=\{q\in\mathbb{R}^{c}\}^{hw}$. Here, for the given patch features $T$, the prototype embedding $Q$ represents a vector-quantized outputs, which indicates the most representative concept $q = \arg\max_{m\in M}\text{cos}(t, m)\in\mathbb{R}^c$ within the concept clusterbook $M$. The segmentation head $S$ comprises a single transformer layer followed by a MLP projection layer only used for training, and we can derive a concept-matched feature $Y=\{y\in\mathbb{R}^{r}\}^{hw}$ for concept fractions in $M$, satisfying $Y=S(Q, T)$ (refer to Appendix \ref{appendix:B}).

\begin{figure}[t!]
\vspace{-12mm}
\begin{algorithm}[H]
\caption{(STEP 2): Enhancing Likelihood of Semantic Groups through Self-Supervised Learning}
\begin{algorithmic}[1]
\Require Head $S$;$\theta_S$, Head-EMA $S_{\text{ema}}$;$\theta_{S_\text{ema}}$, Clusterbook $M$, Distance $\mathcal{D}_M$, Concept Bank $Y_{\text{bank}}$
\For {$X \sim\text{Data}$}
\State $T \gets f(X)$ \Comment{Pre-trained Model Representation}
\State $Q \gets T$ \Comment{Vector Quantization from $M$}
\State $Y, Y_{\text{ema}}\gets S(Q, T), S_{\text{ema}}(Q, T)\quad$ ({\large $\ast$} \text{MLP:} $S(T), S_{\text{ema}}(T)$) \Comment{Segmentation Head Output}
\State $y \sim Y$ \Comment{Anchor Selection (Appendix \ref{appendix:B} for Detail)}
\State $y^+, y^- \sim \{Y_{\text{ema}}, Y_{\text{bank}}\mid y\}$ \Comment{Positive/Negative Selection from $\mathcal{D}_M$ (Appendix \ref{appendix:B} for Detail)}
\State $p \gets \mathbb{E}_{y}\left[\log\mathbb{E}_{y^{+}}\left[\frac{\exp(\text{cos}(y, y^{+}) / \tau)}{\exp(\text{cos}(y, y^{+})/\tau) + \sum_{y^{-}} \exp(\text{cos}(y, y^{-}) / \tau)}\right]\right]$ \Comment{Self-supervised Learning}
\State $\theta_S \gets \text{Increase}(p)$ \Comment{Updating Parameters of Segmentation Head (lr: $0.001$)}
\State $\theta_{S_{\text{ema}}} \gets \lambda\theta_{S_{\text{ema}}} + (1-\lambda)\theta_{S}$ \Comment{Exponential Moving Average ($\lambda:0.99$)}
\State $Y_{\text{bank}} \gets \textbf{R}^{2}(Y_{\text{bank}}, Y_{\text{ema}})$ \Comment{$\textbf{R}^{2}$: \textbf{R}andom Cut $Y_{\text{bank}}$ and \textbf{R}andom Sample $Y_{\text{ema}}$}
\EndFor
\end{algorithmic}
\label{alg:contrastive}
\end{algorithm}
\vspace{-7mm}
\end{figure}

\paragraph{Concept-wise Self-supervised Learning.} 
Using the concept-attended segmentation features, we proceed to enhance the likelihood $p(Y|t', m)$ for effectively clustering pixel-level semantics. To easily handle it, we first re-formulate it as $p(Y|t', m)=\prod_{y\in Y}p(y|t', m)$\footnote{We only utilize the most closest concept at every patch feature point $t$ in $T$. Hence, $p(m|t)$ of Step 1 can be calculated by using sharpening technique: $p(m{=}q|t){=}1$ if it is $q{=}\arg\max_{m\in M} \text{cos}(m,t)$; otherwise, $p(m|t){=}0$. Then, enhancing $\mathop{\mathbb{E}}_{t\in T}\left[p(Y|\text{do}(t) )\right]$ for our main purpose to accomplish unsuperivsed dense prediction can be simplified with increasing $\mathop{\mathbb{E}}_{t\in T}\left[{p(Y|t', m{=}q)}p(t')\right]$. When $p(t')$ is assumed to be uniform distribution, it satisfies $\mathop{\mathbb{E}}_{t\in T}\left[p(Y|\text{do}(t))\right]\uparrow \propto \mathop{\mathbb{E}}_{t\in T}\left[p(Y|t', m{=}q)\right]\uparrow$ so that enhancing the likelihood of semantic groups $Y$ directly leads to increasing causal effect between $T$ and $Y$ even under the presence of $U$.}, recognizing that $Y$ consists of independently learned patch feature points $y\in\mathbb{R}^{r}$. 
However, we cannot directly compute this likelihood as in standard supervised learning, primarily because there are no available pixel annotations. Instead, we substitute the likelihood of unsupervised dense prediction to concept-wise self-supervised learning based on Noise-Contrastive Estimation~\citep{gutmann2010noise}:
\begin{equation}
    p(y\mid t', m)= \mathop{\mathbb{E}}_{y^{+}}\left[\frac{\exp(\text{cos}(y, y^{+}) / \tau)}{\exp(\text{cos}(y, y^{+})/\tau) + \sum_{y^{-}} \exp(\text{cos}(y, y^{-}) / \tau)}\right],
    \label{eqn:likelihood_modification}
\end{equation}
where $y,y^{+},y^{-}$ denote anchor, positive, and negative features, and $\tau$ indicates temperature term.

\paragraph{Positive \& Negative Concept Selection.}
\label{sec:posneg}
When selecting positive and negative concept features for the proposed self-supervised learning, we use a pre-computed distance matrix $\mathcal{D}_M$ that reflects concept-wise similarity between all $k$ concept prototypes such that $\mathcal{D}_M=\text{cos}(M, M)\in\mathbb{R}^{k\times k}$ in concept clusterbook $M$. Specifically, for the given patch feature $t\in\mathbb{R}^{c}$ as an anchor, we can identify the most similar concept $q\in\mathbb{R}^{c}$ and its index: $\text{id}_{q}$ such that $q=\arg\max_{m\in M} \text{cos}(t, m)$. Subsequently, we use the anchor index $\text{id}_{q}$ to access all concept-wise distances for $k$ concept prototypes within $M$ through $\mathcal{D}_{M}[\text{id}_{q}, :]\in\mathbb{R}^{k}$ as pseudo-code-like manner. By using a selection criterion based on the distance $\mathcal{D}_{M}$, we can access concept indices for whole patch features to distinguish positive and negative concept features for the given anchor. That is, once we find patch feature points in $T$ satisfying $\mathcal{D}_M[\text{id}_q, :]>\phi^{+}$ for the given anchor $t$, we designate them as positive concept feature $t^{+}$. Similarly, if they meet the condition $\mathcal{D}_M[\text{id}_q, :]<\phi^{-}$, we categorize them as negative concept feature $t^{-}$. Here, $\phi^{+}$ and $\phi^{-}$ represent the hyper-parameters for positive and negative relaxation, which are both set to $0.3$ and $0.1$, respectively. Note that, we opt for soft relaxation when selecting positive concept features because the main purpose of our unsupervised setup is to group subdivided concept prototypes into the targeted broader-level categories. In this context, a soft positive bound is advantageous as it facilitates a smoother consolidation process. While, we set tight negative relaxation for selecting negative concept features, which aligns with findings in various studies~\citep{khosla2020supervised, kalantidis2020hard, robinson2021contrastive, wang2021exploring} emphasizing that hard negative mining is crucial to advance self-supervised learning.

In the end, after choosing in-batch positive and negative concept features $t^{+}$ and $t^{-}$ for the given anchor $t$, we sample positive segmentation features $y^{+}$ and negative segmentation features $y^{-}$ from the concept-matched $Y=\{y\in\mathbb{R}^{r}\}^{hw}$ within the same spatial location as the selected concept features. Through the concept-wise self-supervised learning in Eq.~(\ref{eqn:likelihood_modification}), we can then guide the segmentation head $S$ to enhance the likelihood of semantic groups $Y$. We re-emphasize that for the given anchor feature (\textit{head}), our goal of USS is the feature consolidation corresponding to positive concept features (\textit{torso, hand, leg, etc.}), and the separation corresponding to negative concept features (\textit{sky, water, board, etc.}), in order to achieve the targeted broader-level semantic groups (\textit{person}).

%%%%%%%%%%%%%%%%%%%%%%%%%%%%%%%%%%%%%%%%%%%%%%%%%%%%%%%%%%%%%%%%%%%%%%%%%%%%%%%%%%%%
\begin{table}[t!]
\vspace{-13mm}
\caption{Comparing quantitative results and applicability to other self-supervised methods on CAUSE.}
\label{table:quantative_table}
\vspace{-2mm}
\centering
\begin{minipage}[t]{0.49\linewidth}
    \centering
\caption*{(a) Experimental results on COCO-Stuff.}
\vspace*{-3mm}
\resizebox{\linewidth}{!}{
\begin{tabular}{llcc}
\Xhline{3\arrayrulewidth} \rule{0pt}{9pt}
Method ($\mathbb{C}=27$)                          & Backbone                                      & mIoU                     & pAcc                     \\ \midrule
IIC~\citep{ji2019invariant}              & ResNet18                                      & 6.7                      & 21.8                     \\
PiCIE~\citep{cho2021picie}               & ResNet18                                      & 14.4                     & 50.0                     \\
SegDiscover~\citep{huang2022segdiscover} & ResNet50                                      & 14.3                     & 40.1                     \\
SlotCon~\citep{wen2022selfsupervised}    & ResNet50                                      & 18.3                     & 42.4                     \\
HSG~\citep{ke2022unsupervised}           & ResNet50                                      & 23.8                     & 57.6                     \\ 
ReCo+~\citep{shin2022reco}               & DeiT-B/8                                       & 32.6                     & 54.1                     \\ \midrule
DINO~\citep{caron2021emerging}           & ViT-S/16                                      & 8.0                      & 22.0                     \\
+ STEGO~\citep{hamilton2022unsupervised} & ViT-S/16                                      & 23.7                     & 52.5                     \\
+ HP~\citep{seong2023leveraging}         & ViT-S/16                                      & 24.3                     & 54.5                     \\
\cdashline{1-4}\noalign{\vskip 0.5ex}
+ \textbf{CAUSE-MLP}                     & ViT-S/16                                      & 25.9                     & 66.3                     \\
+ \textbf{CAUSE-TR}                      & ViT-S/16                                      & \textbf{33.1}            & \textbf{70.4}            \\ \midrule
DINO~\citep{caron2021emerging}           & ViT-S/8                                       & 11.3                     & 28.7                     \\
+ ACSeg~\citep{liacseg}                  & ViT-S/8                                       & 16.4                     & -                        \\
+ TranFGU~\citep{yin2022transfgu}        & ViT-S/8                                       & 17.5                     & 52.7                     \\
+ STEGO~\citep{hamilton2022unsupervised} & ViT-S/8                                       & 24.5                     & 48.3                     \\
+ HP~\citep{seong2023leveraging}         & ViT-S/8                                       & 24.6                     & 57.2                     \\
\cdashline{1-4}\noalign{\vskip 0.5ex}
+ \textbf{CAUSE-MLP}                      & ViT-S/8                                       & 27.9                     & 66.8                     \\
+ \textbf{CAUSE-TR}                       & ViT-S/8                                       & \textbf{32.4}            & \textbf{69.6}            \\ \midrule
DINO~\citep{caron2021emerging}           & ViT-B/8                                       & 13.0                     & 42.4                     \\
+ DINOSAUR~\citep{seitzer2023bridging}   & ViT-B/8                                       & 24.0                     & -                        \\
+ STEGO~\citep{hamilton2022unsupervised} & ViT-B/8                                       & 28.2                     & 56.9                     \\
\cdashline{1-4}\noalign{\vskip 0.5ex}
+ \textbf{CAUSE-MLP}                      & ViT-B/8                                       & 34.3                     & 72.8                     \\
+ \textbf{CAUSE-TR}                       & ViT-B/8                                       & \textbf{41.9}            & \textbf{74.9}            \\ 
\Xhline{3\arrayrulewidth} \rule{0pt}{9pt}                         
\end{tabular}
}
    \vspace{-3mm}
    \centering
\caption*{(c) Self-supervised methods with CAUSE-TR.}
\vspace*{-3mm}
\resizebox{\linewidth}{!}{
\renewcommand{\tabcolsep}{1.22mm}
\begin{tabular}{lllcc}
\Xhline{3\arrayrulewidth} \rule{0pt}{9pt}
Dataset                          & Self-Supervised Methods                             & Backbone     & mIoU     & pAcc            \\\midrule
COCO-Stuff                       & \multirow{3}{*}{DINOv2~\citep{oquab2023dinov2}}     & \multirow{3}{*}{ViT-B/14}      & 45.3     & 78.0           \\ 
Cityscapes                       &                                                     &                                & 29.9     & 89.8           \\
Pascal VOC                       &                                                     &                                & 53.2     & 91.5           \\
\cdashline{1-5}\noalign{\vskip 0.5ex}
COCO-Stuff                       & \multirow{3}{*}{iBOT~\citep{zhou2022image}}         & \multirow{3}{*}{ViT-B/16}      & 39.5     & 73.8           \\ 
Cityscapes                       &                                                     &                                & 23.0     & 89.1           \\
Pascal VOC                       &                                                     &                                & 53.4     & 89.6           \\
\cdashline{1-5}\noalign{\vskip 0.5ex}
COCO-Stuff                       & \multirow{3}{*}{MSN~\citep{assran2022masked}}       & \multirow{3}{*}{ViT-S/16}      & 34.1     & 72.1           \\
Cityscapes                       &                                                     &                                & 21.2     & 89.1           \\
Pascal VOC                       &                                                     &                                & 30.2     & 84.2           \\
\cdashline{1-5}\noalign{\vskip 0.5ex}
COCO-Stuff                       & \multirow{3}{*}{MAE~\citep{he2022masked}}           & \multirow{3}{*}{ViT-B/16}      & 21.5     & 59.1           \\
Cityscapes                       &                                                     &                                & 12.5     & 82.0           \\
Pascal VOC                       &                                                     &                                & 25.8     & 83.7           \\

\Xhline{3\arrayrulewidth} \rule{0pt}{9pt}                         
\end{tabular}
}
\end{minipage}
\begin{minipage}[t]{0.49\linewidth}
    \centering
\caption*{(b) Experimental results on Cityscapes.}
\vspace*{-3mm}
\resizebox{\linewidth}{!}{
\begin{tabular}{llcc}
\Xhline{3\arrayrulewidth} \rule{0pt}{9pt}
Method ($\mathbb{C}=27$)                 & Backbone                  & mIoU            & pAcc           \\\midrule
IIC~\citep{ji2019invariant}              & ResNet18                  & 6.4             & 47.9           \\
PiCIE~\citep{cho2021picie}               & ResNet18                  & 10.3            & 43.0           \\
SegSort~\citep{hwang2019segsort}         & ResNet101                 & 12.3            & 65.5           \\
SegDiscover~\citep{huang2022segdiscover} & ResNet50                  & 24.6            & 81.9           \\
HSG~\citep{ke2022unsupervised}           & ResNet50                  & 32.5            & 86.0           \\
ReCo+~\citep{shin2022reco}               & DeiT-B/8                   & 24.2            & 83.7           \\ \midrule
DINO~\citep{caron2021emerging}           & ViT-S/8                   & 10.9            & 34.5           \\
+ TransFGU~\citep{yin2022transfgu}       & ViT-S/8                   & 16.8            & 77.9           \\
+ HP~\citep{seong2023leveraging}         & ViT-S/8                   & 18.4            & 80.1           \\
\cdashline{1-4}\noalign{\vskip 0.5ex}
+ \textbf{CAUSE-MLP}                      & ViT-S/8                   & 21.7            & 87.7           \\
+ \textbf{CAUSE-TR}                       & ViT-S/8                   & \textbf{24.6}   & \textbf{89.4}  \\ \midrule
DINO~\citep{caron2021emerging}           & ViT-B/8                   & 15.2            & 52.6           \\
+ STEGO~\citep{hamilton2022unsupervised} & ViT-B/8                   & 21.0            & 73.2           \\
+ HP~\citep{seong2023leveraging}         & ViT-B/8                   & 18.4            & 79.5           \\
\cdashline{1-4}\noalign{\vskip 0.5ex}
+ \textbf{CAUSE-MLP}                      & ViT-B/8                   & 25.7            & 90.3           \\
+ \textbf{CAUSE-TR}                       & ViT-B/8                   & \textbf{28.0}   & \textbf{90.8}  \\
\Xhline{3\arrayrulewidth} \rule{0pt}{9pt}                         
\end{tabular}
}
    \vspace{-3mm}
    \centering
\caption*{(d) Experimental results on Pascal VOC 2012.}
\vspace*{-3mm}
\resizebox{\linewidth}{!}{
\begin{tabular}{llc}
\Xhline{3\arrayrulewidth} \rule{0pt}{9pt}
Method ($\mathbb{C}=21$)                    & Backbone       & mIoU             \\\midrule
IIC~\citep{ji2019invariant}                 & ResNet18       & 9.8              \\
SegSort~\citep{hwang2019segsort}            & ResNet101      & 11.7             \\
DenseCL~\citep{wang2021dense}               & ResNet50       & 35.1             \\
HSG~\citep{ke2022unsupervised}              & ResNet50       & 41.9             \\
MaskContrast~\citep{van2021unsupervised}    & ResNet50       & 35.0             \\
MaskDistill~\citep{van2022discovering}      & ResNet50       & 48.9             \\ \midrule
DINO~\citep{caron2021emerging}              & ViT-S/8        & -                \\
+TransFGU~\citep{yin2022transfgu}           & ViT-S/8        & 37.2             \\
+ACSeg~\citep{liacseg}                      & ViT-S/8        & 47.1             \\
\cdashline{1-3}\noalign{\vskip 0.5ex}
+\textbf{CAUSE-MLP}                         & ViT-S/8        & 46.0             \\
+\textbf{CAUSE-TR}                          & ViT-S/8        & \textbf{50.0}    \\ \midrule
DINO~\citep{caron2021emerging}              & ViT-B/8        & -                \\
+DeepSpectral~\citep{melas2022deep}         & ViT-B/8        & 37.2             \\
+DINOSAUR~\citep{seitzer2023bridging}       & ViT-B/8        & 37.2             \\
+Leopart~\citep{ziegler2022self}            & ViT-B/8        & 41.7             \\
+COMUS~\citep{zadaianchuk2023unsupervised}  & ViT-B/8        & 50.0             \\
\cdashline{1-3}\noalign{\vskip 0.5ex}
+\textbf{CAUSE-MLP}                          & ViT-B/8        & 47.9             \\
+\textbf{CAUSE-TR}                           & ViT-B/8        & \textbf{53.3}    \\
\Xhline{3\arrayrulewidth} \rule{0pt}{9pt}                         
\end{tabular}
}
\end{minipage}
\vspace{-7mm}
\end{table}
%%%%%%%%%%%%%%%%%%%%%%%%%%%%%%%%%%%%%%%%%%%%%%%%%%%%%%%%%%%%%%%%%%%%%%%%%%%%%%%%%%%% 

\paragraph{Concept Bank: Out-batch Accumulation.} Unlike image-level self-supervised learning, unsupervised dense prediction requires more intricate pixel-wise comparisons, as discussed in~\citet{zhang2021looking}. To facilitate this, we establish a concept bank, similar to~\citet{he2020momentum} but notably at a pixel-level scale, to accumulate out-batch concept features for additional comparison pairs. Following the same selection criterion as described above, we dynamically sample in-batch features in each training iteration and accumulate them into the concept bank $Y_{\text{bank}}\in\mathbb{R}^{k\times b \times r}$ for continuously utilizing other informative feature from out-batches, where $b$ represents the maximum number of feature points saved for each concept in $M\in\mathbb{R}^{k\times c}$. We incorporate these additional positive and negative concept features into the sets of $y^{+}$ and $y^{-}$ for the concept-wise self-supervised learning. Here, creating a concept bank can be seen as incorporating global views into the pixel-level self-supervised learning beyond local views, which also corresponds to considering all feature representations $T'\in\mathbb{R}^{n\times hw\times c}$ ($n$: total number of images in dataset) for frontdoor adjustment. As a concept bank update strategy, we implement random removal of 50\% of the bank's patch features for each concept prototype, followed by random sampling of 50\% new patch features into the concept bank at every training iteration. In addition, to perform stable self-supervised learning, we employ: (\lowercase\expandafter{\romannumeral1}) using log-probability not to converge to near-zero value due to numerous multiplication of probabilities: $\frac{1}{|Y|}\log p(Y|t', m){=}\frac{1}{|Y|}\log \prod_{y\in Y}p(y|t', m){=}\E_{y\in Y}[\log p(y|t', m)]$, and (\lowercase\expandafter{\romannumeral2}) utilizing exponential moving average (EMA) on teacher-student structure, all of which have been widely used by recent self-supervised learning frameworks such as \citet{grill2020bootstrap, chen2021empirical, caron2021emerging, zhou2022image, assran2022masked}. Please see complete details of Step 2 procedure in Algorithm~\ref{alg:contrastive} and Appendix \ref{appendix:B}.

% Experiment Start
\section{Experiments}
\label{sec:experiment}
\vspace{-1.5mm}
\subsection{Experimental Details}
\vspace{-1.5mm}
\paragraph{Inference.} In inference phase for USS, STEGO~\citep{hamilton2022unsupervised} and HP~\citep{seong2023leveraging} equally perform the following six steps: (a) learning $\mathbb{C}$ cluster centroids~\citep{caron2018deep} from the trained segmentation head output where $\mathbb{C}$ denotes the number of categories in dataset, (b) upsampling segmentation head output to the image resolution, (c) finding the most closest centroid indices to the upsampled output, (d) refining the predicted indices through Fully-connected Conditional Random Field (CRF)~\citep{krahenbuhl2011efficient} with 10 steps, (e) Hungarian Matching~\citep{kuhn1955hungarian} for alignment with CRF indices and true labels, and (f) evaluating mean of intersection over union (mIoU) and pixel accuracy (pAcc). We follow the equal six steps with $S_{\text{ema}}$ of CAUSE.

\paragraph{Implementation.} Following recent works, we adopt DINO as an encoder baseline and freeze it, where the feature dimension $c$ of $T$ depends on the size of ViT: small $(c=384)$ or base $(c=768)$. For hyper-parameter in the clusterbook, the number of concept $k$ in $M$ is set to $2048$ to encompass concept prototypes from pre-trained features as much as possible. During the self-supervised learning, the number of feature accumulation $b$ in concept bank is set to $100$. In addition, output dimension $r$ of segmentation head is set to $90$ based on the dimension analysis~\citep{koenig2023uncovering}. For the segmentation head, we use two variations: (\lowercase\expandafter{\romannumeral1}) \textbf{CAUSE-MLP} with simple MLP layers as in~\citet{hamilton2022unsupervised} and (\lowercase\expandafter{\romannumeral2}) \textbf{CAUSE-TR} with a single layer transformer. Please see details in Appendix \ref{appendix:B}.

\paragraph{Datasets.}
We mainly benchmark CAUSE with three datasets: COCO-Stuff~\citep{caesar2018coco}, Cityscapes~\citep{cordts2016cityscapes}, and Pascal VOC~\citep{everingham2010pascal}. COCO-Stuff is a scene texture segmentation dataset as subset of MS-COCO 2017~\citep{lin2014microsoft} with full pixel annotations of common \textit{Stuff} and \textit{Thing} categories. Cityscapes is an urban street scene parsing dataset with annotations. Following \citet{ji2019invariant, cho2021picie}, we use the curated 27 mid-level categories from label hierarchy for COCO-Stuff and Cityscapes. As an object-centric USS, we follow~\citet{van2022discovering} and report the results of total 21 classes for PASCAL VOC.

%%%%%%%%%%%%%%%%%%%%%%%%%%%%%%%%%%%%%%%%%%%%%%%%%%%%%%%%%%%%%%%%%%%%%%%%%%%%%%%%%%%%
\begin{table}[t!]
\vspace{-13mm}
\centering
\begin{minipage}[t]{0.49\linewidth}
    \centering
\caption{Comparing linear probing performance.}
\label{table:linear_probe}
\vspace*{-3mm}
\resizebox{\linewidth}{!}{
\renewcommand{\tabcolsep}{1mm}
\begin{tabular}{llcccc}
\Xhline{3\arrayrulewidth} \rule{0pt}{9pt}
         &          & \multicolumn{2}{c}{COCO-Stuff} & \multicolumn{2}{c}{Cityscapes} \\ \cmidrule(lr){3-4}\cmidrule(lr){5-6}
Method                                     & Baseline & mIoU           & pAcc          & mIoU           & pAcc          \\ \midrule
DINO~\citep{caron2021emerging}             & ViT-S/8  & 33.9           & 68.6          & 22.8           & 84.6          \\
+HP~\citep{seong2023leveraging}            & ViT-S/8  & 42.7           & 75.6          & 30.6           & 91.2          \\ \cdashline{1-6}\noalign{\vskip 0.5ex}
+\textbf{CAUSE-MLP}                         & ViT-S/8  & 46.4           & 77.3          & 35.2           & 92.1          \\
+\textbf{CAUSE-TR}                          & ViT-S/8  & \textbf{47.2}  & \textbf{78.8} & \textbf{37.2}  & \textbf{93.5} \\
\midrule
DINO~\citep{caron2021emerging}             & ViT-B/8  & 29.4           & 66.8          & 23.0           & 84.2          \\
+STEGO~\citep{hamilton2022unsupervised}    & ViT-B/8  & 41.0           & 76.1          & 26.8           & 90.3          \\ \cdashline{1-6}\noalign{\vskip 0.5ex}
+\textbf{CAUSE-MLP}                         & ViT-B/8  & 48.3           & 79.8          & 38.2           & 93.4          \\ 
+\textbf{CAUSE-TR}                          & ViT-B/8  & \textbf{52.3}  & \textbf{80.1} & \textbf{40.2}  & \textbf{94.5} \\
\Xhline{3\arrayrulewidth} \rule{0pt}{9pt}
\end{tabular}
}

\end{minipage}
\begin{minipage}[t]{0.49\linewidth}
    \centering
\caption{Results of CAUSE with larger categories.}
\label{table:larger_classes}
\vspace*{-3mm}
\resizebox{\linewidth}{!}{
\renewcommand{\tabcolsep}{0.6mm}
\begin{tabular}{cllccc}
\Xhline{3\arrayrulewidth} \rule{0pt}{9pt}
                                                    & Method       & Backbone & mIoU & pAcc \\
\midrule
\multirow{4}{*}{\rotatebox[origin=c]{90}{COCO-81}}  & MaskContrast~\citep{van2021unsupervised} & ResNet50 & 3.7           & 8.8  \\
                                                    & TransFGU~\citep{yin2022transfgu}         & ViT-S/8  & 12.7          & 64.3 \\ \cdashline{2-5}\noalign{\vskip 0.5ex}
                                                    & \textbf{CAUSE-MLP}                       & ViT-S/8  & 19.1          & \textbf{78.8} \\ 
                                                    & \textbf{CAUSE-TR}                        & ViT-S/8  & \textbf{21.2} & 75.2 \\\midrule
\multirow{5}{*}{\rotatebox[origin=c]{90}{COCO-171}} & IIC~\citep{ji2019invariant}              & ResNet50 & 2.2           & 15.7 \\
                                                    & PiCIE~\citep{cho2021picie}               & ResNet50 & 5.6           & 29.8 \\
                                                    & TransFGU~\citep{yin2022transfgu}         & ViT-S/8  & 12.0          & 34.3 \\ \cdashline{2-5}\noalign{\vskip 0.5ex}
                                                    & \textbf{CAUSE-MLP}                       & ViT-S/8  & 10.6          & 44.9 \\
                                                    & \textbf{CAUSE-TR}                        & ViT-S/8  & \textbf{15.2} & \textbf{46.6} \\
\Xhline{3\arrayrulewidth} \rule{0pt}{9pt}
\end{tabular}
}

\end{minipage}
 \vspace{-3mm}
\end{table}
%%%%%%%%%%%%%%%%%%%%%%%%%%%%%%%%%%%%%%%%%%%%%%%%%%%%%%%%%%%%%%%%%%%%%%%%%%%%%%%%%%%%
\begin{figure*}[t!]
\vspace{-3mm}
\centering
\includegraphics[width=0.99\textwidth]{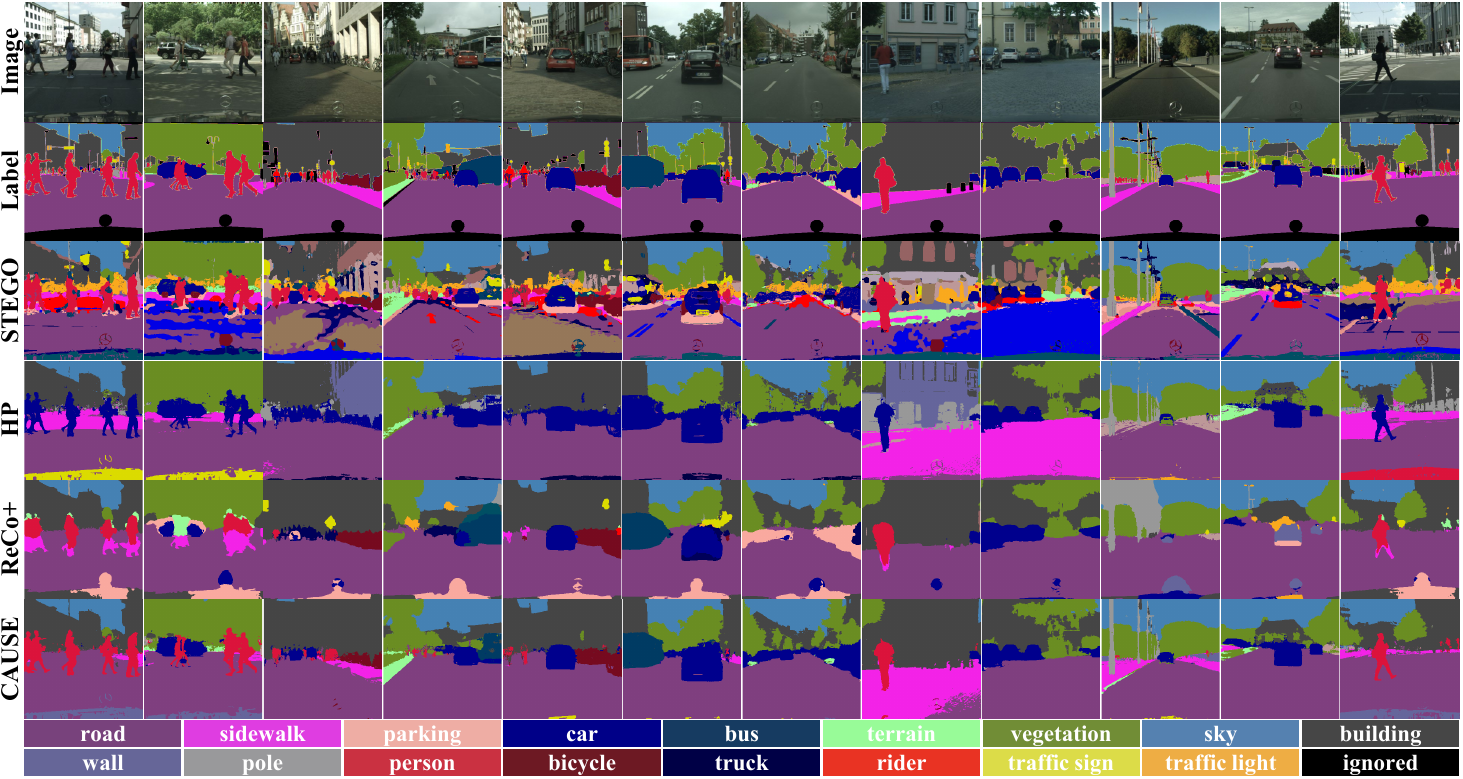}
\vspace*{-2.5mm}
\caption{Qualitative comparison of unsupervised semantic segmentation for Cityscapes dataset.} 
\label{fig:city}
\vspace{-5mm}
\end{figure*}
%%%%%%%%%%%%%%%%%%%%%%%%%%%%%%%%%%%%%%%%%%%%%%%%%%%%%%%%%%%%%%%%%%%%%%%%%%%%%%%%%%%%

\subsection{Validating CAUSE}
\vspace{-1mm}
\paragraph{Quantitative \& Qualitative Results.} We validate CAUSE by comparing with recent USS frameworks using mIoU and pAcc on various datasets. Table~\ref{table:quantative_table} (a) and (b) show CAUSE generally outperforms HSG~\citep{ke2022unsupervised}, TransFGU~\citep{yin2022transfgu}, STEGO~\citep{hamilton2022unsupervised}, HP~\citep{seong2023leveraging}, and ReCo+~\citep{shin2022reco}, and our method achieves state-of-the-art results without any external information. Table~\ref{table:linear_probe} shows another superior quantitative results of CAUSE for linear probing than baselines, which indicates competitive dense representation quality learned in unsupervised manners. Furthermore, Fig.~\ref{fig:1} and Fig.~\ref{fig:city} illustrate CAUSE effectively assembles different level of granularity (\textit{head, torso, hand, leg}, etc.), into one semantically-alike group (\textit{person}). Please see additional qualitative results, analyses, and failure cases in Appendix \ref{appendix:C}.

\paragraph{Applicability to Object-centric Semantic Segmentation.} Preceding works, rooted in object-centric semantic segmentation models~\citep{van2021unsupervised, yin2022transfgu, zadaianchuk2023unsupervised}, initially generate pseudo-labels that differentiate between foreground (objects) and background. This process is typically accomplished by using Mask R-CNN~\citep{he2017mask} and DeepLabv3~\citep{chen2017rethinking}, or saliency maps from DeepUSPS~\citep{nguyen2019deepusps}. In contrast, STEGO and HP abstains from relying on any external information beyond self-supervised knowledge. Therefore, they inherently lack the capability to segment an image into two broad categories: objects and a single background category, making them unsuitable for direct application to object-centric semantic segmentation. However, we highlight that simply adjusting smoother positive relaxation in CAUSE enables to discern background from foreground without any external information. The results of Pascal VOC 2012 is shown in Table~\ref{table:quantative_table}(d), and its figures are illustrated in Appendix \ref{appendix:C}.

%%%%%%%%%%%%%%%%%%%%%%%%%%%%%%%%%%%%%%%%%%%%%%%%%%%%%%%%%%%%%%%%%%%%%%%%%%%%%%%%%%%% 
\begin{figure*}[t!]
\vspace{-13mm}
\centering
\includegraphics[width=\textwidth]{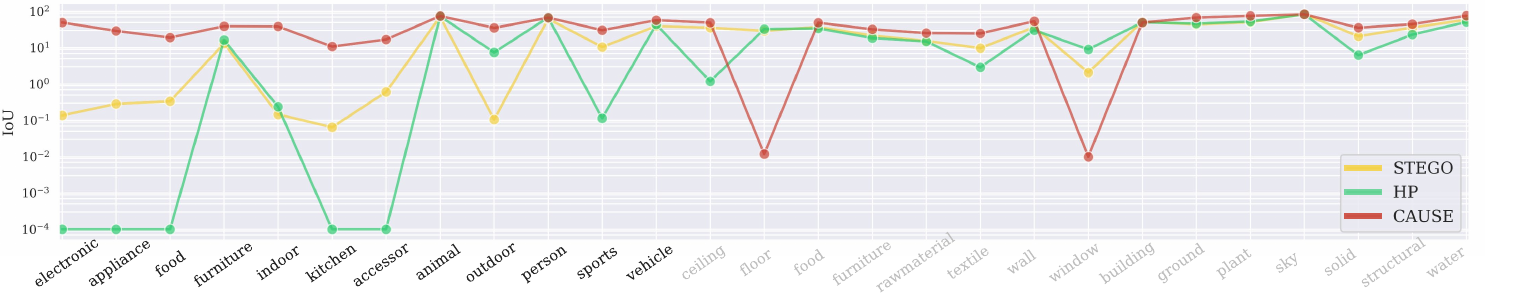}
\vspace*{-7mm}
\begin{flushleft}
    {\hspace{0.23cm} (a) Log scale of IoU results for each categories in COCO-Stuff (Black: \textit{Thing} / \textcolor{lightgray}{Gray: \textit{Stuff}})}
\end{flushleft}
\vspace*{-1mm}
\includegraphics[width=\textwidth]{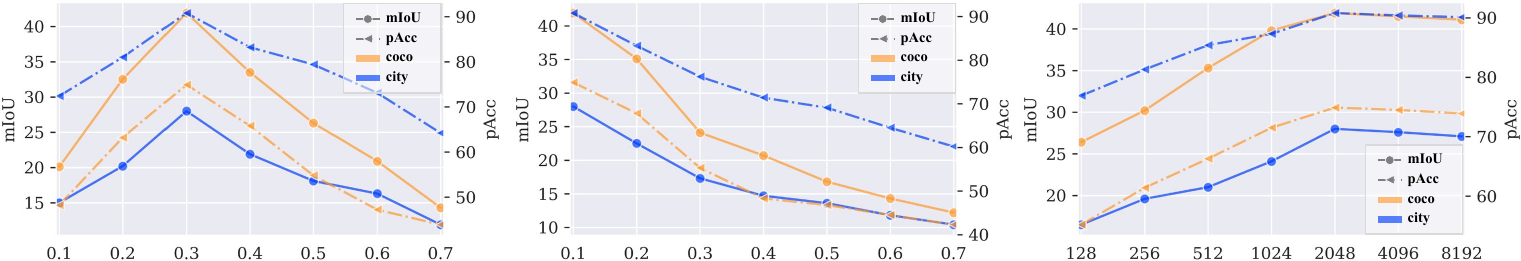}
\vspace*{-6mm}
\begin{flushleft}
    {\hspace{0.35cm} (b) Positive Relaxation $\phi^{+}$ \hspace{0.85cm}(c) Negative Relaxation $\phi^{-}$ \hspace{0.55cm} (d) Concept number $k$ in $M$}
\end{flushleft}	
\vspace*{-0.45cm}
\caption{Additional experimental 
for in-depth analysis and ablation studies of CAUSE-TR.} 
\label{fig:ablation}
\end{figure*}
%%%%%%%%%%%%%%%%%%%%%%%%%%%%%%%%%%%%%%%%%%%%%%%%%%%%%%%%%%%%%%%%%%%%%%%%%%%%%%%%%%%% 
\newcommand{\cmark}{\ding{51}}%
\newcommand{\xmark}{\ding{55}}%
\begin{table}[t!]
\vspace{-3mm}
\caption{Quantitative ablation results by controlling the other three factors of CAUSE-TR on ViT-B/8.}
\vspace*{-3mm}
\label{table:ablation}
\resizebox{\linewidth}{!}{ 
\begin{tabular}{lcccccccccc}
\Xhline{3\arrayrulewidth} \rule{0pt}{9pt} 
$(\%)$                  & \multicolumn{1}{l}{} & \multicolumn{1}{l}{} & \multicolumn{4}{c}{CAUSE-MLP}                                    & \multicolumn{4}{c}{CAUSE-TR}                                     \\
                  \cmidrule(lr){4-7}\cmidrule(lr){8-11}
                  & \multicolumn{2}{c}{}                        & \multicolumn{2}{c}{COCO-Stuff} & \multicolumn{2}{c}{Cityscapes} & \multicolumn{2}{c}{COCO-Stuff} & \multicolumn{2}{c}{Cityscapes} \\
                  \cmidrule(lr){4-5}\cmidrule(lr){6-7}\cmidrule(lr){8-9}\cmidrule(lr){10-11}
Method of Concept Discretization                                       & Bank      & CRF     & mIoU      & pAcc    & mIoU     & pAcc    & mIoU    & pAcc    & mIoU    & pAcc          \\
\midrule
\multirow{4}{*}{Maximizing Modularity~\citep{newman2006modularity}}    & \xmark    & \xmark   & 24.9     & 54.1    & 15.8      & 75.6    & 27.8   & 57.3    & 17.3   & 79.2          \\
\cdashline{2-11}\noalign{\vskip 0.5ex}
                                                                       & \cmark    & \xmark   & \textbf{31.3} & \textbf{69.0} & \textbf{25.3} & \textbf{89.5} & \textbf{39.5}  & \textbf{72.5} & \textbf{28.8}  & \textbf{90.7}\\
\cmidrule{2-11}
                                                                       & \xmark    & \cmark   & 27.5     & 57.9    & 17.3      & 78.8     & 30.3   & 60.1    & 19.6   & 82.1          \\
\cdashline{2-11}\noalign{\vskip 0.5ex}
                                                                       & \cmark     & \cmark  & \textbf{34.3} & \textbf{72.8} & \textbf{25.7} & \textbf{90.3} & \textbf{41.9}  & \textbf{74.9} & \textbf{28.0}  & \textbf{90.8} \\
\midrule
K-Means++~\citep{arthur2007k}                                          & \cmark     & \cmark  & 27.8      & 64.7    & 18.9      & 81.3    & 33.7    & 62.7   & 20.4   & 83.2          \\
\cdashline{2-11}\noalign{\vskip 0.5ex}
Spectral Clustering~\citep{von2007tutorial}                            & \cmark     & \cmark  & 30.7      & 65.1    & 20.8      & 83.5    & 35.9    & 66.7   & 22.8   & 84.1          \\
\cdashline{2-11}\noalign{\vskip 0.5ex}
Agglomerative Clustering~\citep{mullner2011modern}                     & \cmark     & \cmark  & 31.4      & 67.9    & 22.2      & 84.0    & 37.7    & 68.1   & 24.5   & 86.3          \\
\cdashline{2-11}\noalign{\vskip 0.5ex}
Ward-Hierarchical Clustering~\citep{murtagh2014ward}                   & \cmark     & \cmark  & 31.8      & 67.5    & 22.9      & 84.7    & 37.5    & 68.2   & 24.7   & 87.0          \\
\Xhline{3\arrayrulewidth} \rule{0pt}{9pt} 
\end{tabular}
}
\vspace{-7mm}
\end{table}
%%%%%%%%%%%%%%%%%%%%%%%%%%%%%%%%%%%%%%%%%%%%%%%%%%%%%%%%%%%%%%%%%%%%%%%%%%%%%%%%%%%% 

\paragraph{Generalization Capability} We first incorporate alternative self-supervised methods as our baseline, replacing DINO~\citep{caron2021emerging}. In Table~\ref{table:quantative_table}(c), we present an overview of adaptability in CAUSE across DINOv2~\citep{oquab2023dinov2}, iBOT~\citep{zhou2022image}, MSN~\citep{assran2022masked}, and MAE~\citep{he2022masked}. Furthermore, we extend the number of clusters in CAUSE by utilizing MS-COCO 2017~\citep{lin2014microsoft}, which comprises 80 object categories and one background category: (object-centric) COCO-81, and 171 categories encompassing both \textit{Stuff} and \textit{Thing} categories: COCO-171. Note that, positive $\phi^+$ relaxation is set to $0.4$ and $0.55$ respectively. Table~\ref{table:larger_classes} highlights CAUSE retains superior performances for USS even with larger categories. Especially, TransFGU~\citep{yin2022transfgu} used Grad-CAM~\citep{selvaraju2017grad} for generating category-specific pseudo-labels, thereby keeping consistent mIoU performance compared with COCO-81 and COCO-171. Nonetheless, CAUSE has a great advantage to pAcc especially in COCO-171 without any external information.

\paragraph{Categorical Analysis.} To demonstrate that CAUSE can effectively address the targeted level of semantic grouping, we closely examine IoU results for each category. By validating the IoU results on a logarithmic scale in Fig.~\ref{fig:ablation}(a), we can observe that STEGO and HP struggle with segmenting \textit{Thing} categories in COCO-Stuff, which demands fine-grained discrimination among concepts within complex scenes. In contrast, CAUSE consistently exhibits superior capability in segmenting concepts across most categories. These results are largely attributed to the causal design aspects, including the construction of the concept clusterbook and concept-wise self-supervised learning among concept prototypes. Beyond the quantitative results, it is important to highlight again that CAUSE exhibits significantly improved visual quality in achieving targeted level of semantic groupings than baselines as in Fig.~\ref{fig:1} and Fig.~\ref{fig:city}. We include further discussions and limitations in Appendix \ref{appendix:D}.

\paragraph{Ablation Studies.} We conduct ablation studies on six factors of CAUSE to identify where the effectiveness comes from as in Fig.~\ref{fig:ablation} and Table~\ref{table:ablation}: (\lowercase\expandafter{\romannumeral1}) positive $\phi^{+}$ and (\lowercase\expandafter{\romannumeral2}) negative relaxation $\phi^{-}$, (\lowercase\expandafter{\romannumeral3}) the number of concepts $k$ in $M$, (\lowercase\expandafter{\romannumeral4}) the effects of concept bank $Y_{\text{bank}}$ and (\lowercase\expandafter{\romannumeral5}) fully-connected CRF, and (\lowercase\expandafter{\romannumeral6}) discretizing methods for concept clusterbook $M$. Through the empirical results, we first observe the appropriate relaxation parameter plays a crucial role in determining the quality of self-supervised learning. Furthermore, unlike semantic representation-level pre-training~\citep{bao2022beit}, we find that the number of discretized concepts saturates after reaching $2048$ for clustering. We also highlight the effects of concept bank, CRF, and modularity maximization for effective USS.

% Conclustion Start
\section{Conclusion}
\label{sec:4}
In this work, we propose a novel framework called CAusal Unsupervised Semantic sEgmentation (CAUSE) to address the indeterminate clustering targets that exist in unsupervised semantic segmentation tasks. By employing frontdoor adjustment, we construct the concept clusterbook as a mediator and utilize the concept prototypes for semantic grouping through concept-wise self-supervised learning. Extensive experiments demonstrate the effectiveness of CAUSE, resulting in state-of-the-art performance in unsupervised semantic segmentation. Our findings bridge causal perspectives into the unsupervised prediction, and improve segmentation quality without any pixel annotations.

% Acknowledgements

% Reference
\bibliography{iclr2024_conference}

\begin{thebibliography}{88}
\providecommand{\natexlab}[1]{#1}
\providecommand{\url}[1]{\texttt{#1}}
\expandafter\ifx\csname urlstyle\endcsname\relax
  \providecommand{\doi}[1]{doi: #1}\else
  \providecommand{\doi}{doi: \begingroup \urlstyle{rm}\Url}\fi

\bibitem[Agarwal et~al.(2020)Agarwal, Shetty, and Fritz]{agarwal2020towards}
Vedika Agarwal, Rakshith Shetty, and Mario Fritz.
\newblock Towards causal vqa: Revealing and reducing spurious correlations by invariant and covariant semantic editing.
\newblock In \emph{Proceedings of the IEEE/CVF Conference on Computer Vision and Pattern Recognition}, pp.\  9690--9698, 2020.

\bibitem[Arthur \& Vassilvitskii(2007)Arthur and Vassilvitskii]{arthur2007k}
David Arthur and Sergei Vassilvitskii.
\newblock K-means++ the advantages of careful seeding.
\newblock In \emph{Proceedings of the eighteenth annual ACM-SIAM symposium on Discrete algorithms}, pp.\  1027--1035, 2007.

\bibitem[Assent(2012)]{assent2012clustering}
Ira Assent.
\newblock Clustering high dimensional data.
\newblock \emph{Wiley Interdisciplinary Reviews: Data Mining and Knowledge Discovery}, 2\penalty0 (4):\penalty0 340--350, 2012.

\bibitem[Assran et~al.(2022)Assran, Caron, Misra, Bojanowski, Bordes, Vincent, Joulin, Rabbat, and Ballas]{assran2022masked}
Mahmoud Assran, Mathilde Caron, Ishan Misra, Piotr Bojanowski, Florian Bordes, Pascal Vincent, Armand Joulin, Mike Rabbat, and Nicolas Ballas.
\newblock Masked siamese networks for label-efficient learning.
\newblock In \emph{European Conference on Computer Vision}, pp.\  456--473. Springer, 2022.

\bibitem[Bao et~al.(2022)Bao, Dong, Piao, and Wei]{bao2022beit}
Hangbo Bao, Li~Dong, Songhao Piao, and Furu Wei.
\newblock {BE}it: {BERT} pre-training of image transformers.
\newblock In \emph{International Conference on Learning Representations}, 2022.

\bibitem[Caesar et~al.(2018)Caesar, Uijlings, and Ferrari]{caesar2018coco}
Holger Caesar, Jasper Uijlings, and Vittorio Ferrari.
\newblock Coco-stuff: Thing and stuff classes in context.
\newblock In \emph{Proceedings of the IEEE/CVF Conference on Computer Vision and Pattern Recognition}, pp.\  1209--1218, 2018.

\bibitem[Carion et~al.(2020)Carion, Massa, Synnaeve, Usunier, Kirillov, and Zagoruyko]{carion2020end}
Nicolas Carion, Francisco Massa, Gabriel Synnaeve, Nicolas Usunier, Alexander Kirillov, and Sergey Zagoruyko.
\newblock End-to-end object detection with transformers.
\newblock In \emph{European Conference on Computer Vision}, pp.\  213--229. Springer, 2020.

\bibitem[Caron et~al.(2018)Caron, Bojanowski, Joulin, and Douze]{caron2018deep}
Mathilde Caron, Piotr Bojanowski, Armand Joulin, and Matthijs Douze.
\newblock Deep clustering for unsupervised learning of visual features.
\newblock In \emph{European Conference on Computer Vision}, pp.\  132--149, 2018.

\bibitem[Caron et~al.(2021)Caron, Touvron, Misra, J{\'e}gou, Mairal, Bojanowski, and Joulin]{caron2021emerging}
Mathilde Caron, Hugo Touvron, Ishan Misra, Herv{\'e} J{\'e}gou, Julien Mairal, Piotr Bojanowski, and Armand Joulin.
\newblock Emerging properties in self-supervised vision transformers.
\newblock In \emph{Proceedings of the IEEE/CVF International Conference on Computer Vision}, pp.\  9650--9660, 2021.

\bibitem[Chen et~al.(2017)Chen, Papandreou, Schroff, and Adam]{chen2017rethinking}
Liang-Chieh Chen, George Papandreou, Florian Schroff, and Hartwig Adam.
\newblock Rethinking atrous convolution for semantic image segmentation.
\newblock \emph{arXiv preprint arXiv:1706.05587}, 2017.

\bibitem[Chen et~al.(2021)Chen, Xie, and He]{chen2021empirical}
Xinlei Chen, Saining Xie, and Kaiming He.
\newblock An empirical study of training self-supervised vision transformers.
\newblock In \emph{Proceedings of the IEEE/CVF International Conference on Computer Vision}, pp.\  9640--9649, 2021.

\bibitem[Cho et~al.(2021)Cho, Mall, Bala, and Hariharan]{cho2021picie}
Jang~Hyun Cho, Utkarsh Mall, Kavita Bala, and Bharath Hariharan.
\newblock Picie: Unsupervised semantic segmentation using invariance and equivariance in clustering.
\newblock In \emph{Proceedings of the IEEE/CVF Conference on Computer Vision and Pattern Recognition}, pp.\  16794--16804, 2021.

\bibitem[Cordts et~al.(2016)Cordts, Omran, Ramos, Rehfeld, Enzweiler, Benenson, Franke, Roth, and Schiele]{cordts2016cityscapes}
Marius Cordts, Mohamed Omran, Sebastian Ramos, Timo Rehfeld, Markus Enzweiler, Rodrigo Benenson, Uwe Franke, Stefan Roth, and Bernt Schiele.
\newblock The cityscapes dataset for semantic urban scene understanding.
\newblock In \emph{Proceedings of the IEEE/CVF Conference on Computer Vision and Pattern Recognition}, pp.\  3213--3223, 2016.

\bibitem[Dai et~al.(2015)Dai, He, and Sun]{dai2015boxsup}
Jifeng Dai, Kaiming He, and Jian Sun.
\newblock Boxsup: Exploiting bounding boxes to supervise convolutional networks for semantic segmentation.
\newblock In \emph{Proceedings of the IEEE/CVF Conference on Computer Vision and Pattern Recognition}, pp.\  1635--1643, 2015.

\bibitem[Dosovitskiy et~al.(2020)Dosovitskiy, Beyer, Kolesnikov, Weissenborn, Zhai, Unterthiner, Dehghani, Minderer, Heigold, Gelly, et~al.]{dosovitskiy2020image}
Alexey Dosovitskiy, Lucas Beyer, Alexander Kolesnikov, Dirk Weissenborn, Xiaohua Zhai, Thomas Unterthiner, Mostafa Dehghani, Matthias Minderer, Georg Heigold, Sylvain Gelly, et~al.
\newblock An image is worth 16x16 words: Transformers for image recognition at scale.
\newblock In \emph{International Conference on Learning Representations}, 2020.

\bibitem[Esser et~al.(2021)Esser, Rombach, and Ommer]{esser2021taming}
Patrick Esser, Robin Rombach, and Bjorn Ommer.
\newblock Taming transformers for high-resolution image synthesis.
\newblock In \emph{Proceedings of the IEEE/CVF Conference on Computer Vision and Pattern Recognition}, pp.\  12873--12883, 2021.

\bibitem[Everingham et~al.(2010)Everingham, Van~Gool, Williams, Winn, and Zisserman]{everingham2010pascal}
Mark Everingham, Luc Van~Gool, Christopher~KI Williams, John Winn, and Andrew Zisserman.
\newblock The pascal visual object classes (voc) challenge.
\newblock \emph{International journal of computer vision}, 88:\penalty0 303--338, 2010.

\bibitem[Grill et~al.(2020)Grill, Strub, Altch{\'e}, Tallec, Richemond, Buchatskaya, Doersch, Avila~Pires, Guo, Gheshlaghi~Azar, et~al.]{grill2020bootstrap}
Jean-Bastien Grill, Florian Strub, Florent Altch{\'e}, Corentin Tallec, Pierre Richemond, Elena Buchatskaya, Carl Doersch, Bernardo Avila~Pires, Zhaohan Guo, Mohammad Gheshlaghi~Azar, et~al.
\newblock Bootstrap your own latent-a new approach to self-supervised learning.
\newblock \emph{Advances in Neural Information Processing Systems}, 33:\penalty0 21271--21284, 2020.

\bibitem[Gutmann \& Hyv{\"a}rinen(2010)Gutmann and Hyv{\"a}rinen]{gutmann2010noise}
Michael Gutmann and Aapo Hyv{\"a}rinen.
\newblock Noise-contrastive estimation: A new estimation principle for unnormalized statistical models.
\newblock In \emph{Proceedings of the thirteenth international conference on artificial intelligence and statistics}, pp.\  297--304. JMLR Workshop and Conference Proceedings, 2010.

\bibitem[Hamilton et~al.(2022)Hamilton, Zhang, Hariharan, Snavely, and Freeman]{hamilton2022unsupervised}
Mark Hamilton, Zhoutong Zhang, Bharath Hariharan, Noah Snavely, and William~T. Freeman.
\newblock Unsupervised semantic segmentation by distilling feature correspondences.
\newblock In \emph{International Conference on Learning Representations}, 2022.

\bibitem[He et~al.(2016)He, Zhang, Ren, and Sun]{resnet}
Kaiming He, Xiangyu Zhang, Shaoqing Ren, and Jian Sun.
\newblock Deep residual learning for image recognition.
\newblock In \emph{Proceedings of the IEEE/CVF Conference on Computer Vision and Pattern Recognition}, pp.\  770--778, 2016.

\bibitem[He et~al.(2017)He, Gkioxari, Doll{\'a}r, and Girshick]{he2017mask}
Kaiming He, Georgia Gkioxari, Piotr Doll{\'a}r, and Ross Girshick.
\newblock Mask r-cnn.
\newblock In \emph{Proceedings of the IEEE International Conference on Computer Vision}, pp.\  2961--2969, 2017.

\bibitem[He et~al.(2020)He, Fan, Wu, Xie, and Girshick]{he2020momentum}
Kaiming He, Haoqi Fan, Yuxin Wu, Saining Xie, and Ross Girshick.
\newblock Momentum contrast for unsupervised visual representation learning.
\newblock In \emph{Proceedings of the IEEE/CVF Conference on Computer Vision and Pattern Recognition}, pp.\  9729--9738, 2020.

\bibitem[He et~al.(2022)He, Chen, Xie, Li, Doll{\'a}r, and Girshick]{he2022masked}
Kaiming He, Xinlei Chen, Saining Xie, Yanghao Li, Piotr Doll{\'a}r, and Ross Girshick.
\newblock Masked autoencoders are scalable vision learners.
\newblock In \emph{Proceedings of the IEEE/CVF Conference on Computer Vision and Pattern Recognition}, pp.\  16000--16009, 2022.

\bibitem[Huang et~al.(2022)Huang, Chen, and Rudin]{huang2022segdiscover}
Haiyang Huang, Zhi Chen, and Cynthia Rudin.
\newblock Segdiscover: Visual concept discovery via unsupervised semantic segmentation.
\newblock \emph{arXiv preprint arXiv:2204.10926}, 2022.

\bibitem[Hwang et~al.(2019)Hwang, Yu, Shi, Collins, Yang, Zhang, and Chen]{hwang2019segsort}
Jyh-Jing Hwang, Stella~X Yu, Jianbo Shi, Maxwell~D Collins, Tien-Ju Yang, Xiao Zhang, and Liang-Chieh Chen.
\newblock Segsort: Segmentation by discriminative sorting of segments.
\newblock In \emph{Proceedings of the IEEE/CVF International Conference on Computer Vision}, pp.\  7334--7344, 2019.

\bibitem[Ji et~al.(2019)Ji, Henriques, and Vedaldi]{ji2019invariant}
Xu~Ji, Joao~F Henriques, and Andrea Vedaldi.
\newblock Invariant information clustering for unsupervised image classification and segmentation.
\newblock In \emph{Proceedings of the IEEE/CVF International Conference on Computer Vision}, pp.\  9865--9874, 2019.

\bibitem[Kalantidis et~al.(2020)Kalantidis, Sariyildiz, Pion, Weinzaepfel, and Larlus]{kalantidis2020hard}
Yannis Kalantidis, Mert~Bulent Sariyildiz, Noe Pion, Philippe Weinzaepfel, and Diane Larlus.
\newblock Hard negative mixing for contrastive learning.
\newblock \emph{Advances in Neural Information Processing Systems}, 33:\penalty0 21798--21809, 2020.

\bibitem[Ke et~al.(2022)Ke, Hwang, Guo, Wang, and Yu]{ke2022unsupervised}
Tsung-Wei Ke, Jyh-Jing Hwang, Yunhui Guo, Xudong Wang, and Stella~X Yu.
\newblock Unsupervised hierarchical semantic segmentation with multiview cosegmentation and clustering transformers.
\newblock In \emph{Proceedings of the IEEE/CVF Conference on Computer Vision and Pattern Recognition}, pp.\  2571--2581, 2022.

\bibitem[Khoreva et~al.(2017)Khoreva, Benenson, Hosang, Hein, and Schiele]{khoreva2017simple}
Anna Khoreva, Rodrigo Benenson, Jan Hosang, Matthias Hein, and Bernt Schiele.
\newblock Simple does it: Weakly supervised instance and semantic segmentation.
\newblock In \emph{Proceedings of the IEEE/CVF Conference on Computer Vision and Pattern Recognition}, pp.\  876--885, 2017.

\bibitem[Khosla et~al.(2020)Khosla, Teterwak, Wang, Sarna, Tian, Isola, Maschinot, Liu, and Krishnan]{khosla2020supervised}
Prannay Khosla, Piotr Teterwak, Chen Wang, Aaron Sarna, Yonglong Tian, Phillip Isola, Aaron Maschinot, Ce~Liu, and Dilip Krishnan.
\newblock Supervised contrastive learning.
\newblock \emph{Advances in Neural Information Processing Systems}, 33:\penalty0 18661--18673, 2020.

\bibitem[Kim et~al.(2023{\natexlab{a}})Kim, Kim, Cho, Luo, and Hong]{kim2023universal}
Donggyun Kim, Jinwoo Kim, Seongwoong Cho, Chong Luo, and Seunghoon Hong.
\newblock Universal few-shot learning of dense prediction tasks with visual token matching.
\newblock In \emph{International Conference on Learning Representations}, 2023{\natexlab{a}}.

\bibitem[Kim et~al.(2023{\natexlab{b}})Kim, Lee, and Ro]{kim2023demystifying}
Junho Kim, Byung-Kwan Lee, and Yong~Man Ro.
\newblock Demystifying causal features on adversarial examples and causal inoculation for robust network by adversarial instrumental variable regression.
\newblock In \emph{Proceedings of the IEEE/CVF Conference on Computer Vision and Pattern Recognition}, pp.\  12302--12312, 2023{\natexlab{b}}.

\bibitem[Kingma \& Ba(2015)Kingma and Ba]{kingma2014adam}
Diederik Kingma and Jimmy Ba.
\newblock Adam: A method for stochastic optimization.
\newblock In \emph{International Conference on Learning Representations)}, San Diega, CA, USA, 2015.

\bibitem[Kirillov et~al.(2019)Kirillov, Girshick, He, and Doll{\'a}r]{kirillov2019panoptic}
Alexander Kirillov, Ross Girshick, Kaiming He, and Piotr Doll{\'a}r.
\newblock Panoptic feature pyramid networks.
\newblock In \emph{Proceedings of the IEEE/CVF Conference on Computer Vision and Pattern Recognition}, pp.\  6399--6408, 2019.

\bibitem[Koenig et~al.(2023)Koenig, Schambach, and Otterbach]{koenig2023uncovering}
Alexander Koenig, Maximilian Schambach, and Johannes Otterbach.
\newblock Uncovering the inner workings of stego for safe unsupervised semantic segmentation.
\newblock In \emph{Proceedings of the IEEE/CVF Conference on Computer Vision and Pattern Recognition}, pp.\  3788--3797, 2023.

\bibitem[Kr{\"a}henb{\"u}hl \& Koltun(2011)Kr{\"a}henb{\"u}hl and Koltun]{krahenbuhl2011efficient}
Philipp Kr{\"a}henb{\"u}hl and Vladlen Koltun.
\newblock Efficient inference in fully connected crfs with gaussian edge potentials.
\newblock \emph{Advances in Neural Information Processing Systems}, 24, 2011.

\bibitem[Kuhn(1955)]{kuhn1955hungarian}
Harold~W Kuhn.
\newblock The hungarian method for the assignment problem.
\newblock \emph{Naval research logistics quarterly}, 2\penalty0 (1-2):\penalty0 83--97, 1955.

\bibitem[Lee et~al.(2023)Lee, Kim, and Ro]{lee2023mitigating}
Byung-Kwan Lee, Junho Kim, and Yong~Man Ro.
\newblock Mitigating adversarial vulnerability through causal parameter estimation by adversarial double machine learning.
\newblock In \emph{Proceedings of the IEEE/CVF International Conference on Computer Vision}, pp.\  4499--4509, October 2023.

\bibitem[Li et~al.(2023)Li, Wang, Cheng, Yu, Zhao, Song, Liu, Yuan, and Chen]{liacseg}
Kehan Li, Zhennan Wang, Zesen Cheng, Runyi Yu, Yian Zhao, Guoli Song, Chang Liu, Li~Yuan, and Jie Chen.
\newblock Acseg: Adaptive conceptualization for unsupervised semantic segmentation.
\newblock In \emph{Proceedings of the IEEE/CVF Conference on Computer Vision and Pattern Recognition}, 2023.

\bibitem[Lin et~al.(2016)Lin, Dai, Jia, He, and Sun]{lin2016scribblesup}
Di~Lin, Jifeng Dai, Jiaya Jia, Kaiming He, and Jian Sun.
\newblock Scribblesup: Scribble-supervised convolutional networks for semantic segmentation.
\newblock In \emph{Proceedings of the IEEE/CVF Conference on Computer Vision and Pattern Recognition}, pp.\  3159--3167, 2016.

\bibitem[Lin et~al.(2014)Lin, Maire, Belongie, Hays, Perona, Ramanan, Doll{\'a}r, and Zitnick]{lin2014microsoft}
Tsung-Yi Lin, Michael Maire, Serge Belongie, James Hays, Pietro Perona, Deva Ramanan, Piotr Doll{\'a}r, and C~Lawrence Zitnick.
\newblock Microsoft coco: Common objects in context.
\newblock In \emph{European Conference on Computer Vision}, pp.\  740--755. Springer, 2014.

\bibitem[Liu et~al.(2022)Liu, Wang, Yang, Zhou, Yao, Shao, and Zhao]{liu2022show}
Bing Liu, Dong Wang, Xu~Yang, Yong Zhou, Rui Yao, Zhiwen Shao, and Jiaqi Zhao.
\newblock Show, deconfound and tell: Image captioning with causal inference.
\newblock In \emph{Proceedings of the IEEE/CVF Conference on Computer Vision and Pattern Recognition}, pp.\  18041--18050, 2022.

\bibitem[Lv et~al.(2022)Lv, Liang, Li, Zang, Liu, Wang, and Liu]{lv2022causality}
Fangrui Lv, Jian Liang, Shuang Li, Bin Zang, Chi~Harold Liu, Ziteng Wang, and Di~Liu.
\newblock Causality inspired representation learning for domain generalization.
\newblock In \emph{Proceedings of the IEEE/CVF Conference on Computer Vision and Pattern Recognition}, pp.\  8046--8056, 2022.

\bibitem[Melas-Kyriazi et~al.(2022)Melas-Kyriazi, Rupprecht, Laina, and Vedaldi]{melas2022deep}
Luke Melas-Kyriazi, Christian Rupprecht, Iro Laina, and Andrea Vedaldi.
\newblock Deep spectral methods: A surprisingly strong baseline for unsupervised semantic segmentation and localization.
\newblock In \emph{Proceedings of the IEEE/CVF Conference on Computer Vision and Pattern Recognition}, pp.\  8364--8375, 2022.

\bibitem[M{\"u}llner(2011)]{mullner2011modern}
Daniel M{\"u}llner.
\newblock Modern hierarchical, agglomerative clustering algorithms.
\newblock \emph{arXiv preprint arXiv:1109.2378}, 2011.

\bibitem[Murtagh \& Legendre(2014)Murtagh and Legendre]{murtagh2014ward}
Fionn Murtagh and Pierre Legendre.
\newblock Ward’s hierarchical agglomerative clustering method: which algorithms implement ward’s criterion?
\newblock \emph{Journal of classification}, 31:\penalty0 274--295, 2014.

\bibitem[Newman(2006)]{newman2006modularity}
Mark~EJ Newman.
\newblock Modularity and community structure in networks.
\newblock \emph{Proceedings of the National Academy of Sciencess}, 103\penalty0 (23):\penalty0 8577--8582, 2006.

\bibitem[Nguyen et~al.(2019)Nguyen, Dax, Mummadi, Ngo, Nguyen, Lou, and Brox]{nguyen2019deepusps}
Tam Nguyen, Maximilian Dax, Chaithanya~Kumar Mummadi, Nhung Ngo, Thi Hoai~Phuong Nguyen, Zhongyu Lou, and Thomas Brox.
\newblock Deepusps: Deep robust unsupervised saliency prediction via self-supervision.
\newblock \emph{Advances in Neural Information Processing Systems}, 32, 2019.

\bibitem[Niu et~al.(2021)Niu, Tang, Zhang, Lu, Hua, and Wen]{niu2021counterfactual}
Yulei Niu, Kaihua Tang, Hanwang Zhang, Zhiwu Lu, Xian-Sheng Hua, and Ji-Rong Wen.
\newblock Counterfactual vqa: A cause-effect look at language bias.
\newblock In \emph{Proceedings of the IEEE/CVF Conference on Computer Vision and Pattern Recognition}, pp.\  12700--12710, 2021.

\bibitem[Oord et~al.(2018)Oord, Li, and Vinyals]{oord2018representation}
Aaron van~den Oord, Yazhe Li, and Oriol Vinyals.
\newblock Representation learning with contrastive predictive coding.
\newblock \emph{arXiv preprint arXiv:1807.03748}, 2018.

\bibitem[Oquab et~al.(2023)Oquab, Darcet, Moutakanni, Vo, Szafraniec, Khalidov, Fernandez, Haziza, Massa, El-Nouby, et~al.]{oquab2023dinov2}
Maxime Oquab, Timoth{\'e}e Darcet, Th{\'e}o Moutakanni, Huy Vo, Marc Szafraniec, Vasil Khalidov, Pierre Fernandez, Daniel Haziza, Francisco Massa, Alaaeldin El-Nouby, et~al.
\newblock Dinov2: Learning robust visual features without supervision.
\newblock \emph{arXiv preprint arXiv:2304.07193}, 2023.

\bibitem[Pearl(1993)]{pearl1993bayesian}
Judea Pearl.
\newblock [bayesian analysis in expert systems]: comment: graphical models, causality and intervention.
\newblock \emph{Statistical Science}, 8\penalty0 (3):\penalty0 266--269, 1993.

\bibitem[Pearl(1995)]{pearl1995causal}
Judea Pearl.
\newblock Causal diagrams for empirical research.
\newblock \emph{Biometrika}, 82\penalty0 (4):\penalty0 669--688, 1995.

\bibitem[Pearl(2009)]{pearl2009causality}
Judea Pearl.
\newblock \emph{Causality}.
\newblock Cambridge university press, 2009.

\bibitem[Ranftl et~al.(2021)Ranftl, Bochkovskiy, and Koltun]{ranftl2021vision}
Ren{\'e} Ranftl, Alexey Bochkovskiy, and Vladlen Koltun.
\newblock Vision transformers for dense prediction.
\newblock In \emph{Proceedings of the IEEE/CVF International Conference on Computer Vision}, pp.\  12179--12188, 2021.

\bibitem[Robinson et~al.(2021)Robinson, Chuang, Sra, and Jegelka]{robinson2021contrastive}
Joshua~David Robinson, Ching-Yao Chuang, Suvrit Sra, and Stefanie Jegelka.
\newblock Contrastive learning with hard negative samples.
\newblock In \emph{International Conference on Learning Representations}, 2021.

\bibitem[Sch{\"o}lkopf et~al.(2021)Sch{\"o}lkopf, Locatello, Bauer, Ke, Kalchbrenner, Goyal, and Bengio]{scholkopf2021toward}
Bernhard Sch{\"o}lkopf, Francesco Locatello, Stefan Bauer, Nan~Rosemary Ke, Nal Kalchbrenner, Anirudh Goyal, and Yoshua Bengio.
\newblock Toward causal representation learning.
\newblock \emph{Proceedings of the IEEE}, 109\penalty0 (5):\penalty0 612--634, 2021.

\bibitem[Seitzer et~al.(2023)Seitzer, Horn, Zadaianchuk, Zietlow, Xiao, Simon-Gabriel, He, Zhang, Sch{\"o}lkopf, Brox, and Locatello]{seitzer2023bridging}
Maximilian Seitzer, Max Horn, Andrii Zadaianchuk, Dominik Zietlow, Tianjun Xiao, Carl-Johann Simon-Gabriel, Tong He, Zheng Zhang, Bernhard Sch{\"o}lkopf, Thomas Brox, and Francesco Locatello.
\newblock Bridging the gap to real-world object-centric learning.
\newblock In \emph{International Conference on Learning Representations}, 2023.

\bibitem[Selvaraju et~al.(2017)Selvaraju, Cogswell, Das, Vedantam, Parikh, and Batra]{selvaraju2017grad}
Ramprasaath~R Selvaraju, Michael Cogswell, Abhishek Das, Ramakrishna Vedantam, Devi Parikh, and Dhruv Batra.
\newblock Grad-cam: Visual explanations from deep networks via gradient-based localization.
\newblock In \emph{Proceedings of the IEEE International Conference on Computer Vision}, pp.\  618--626, 2017.

\bibitem[Seong et~al.(2023)Seong, Moon, Lee, and Heo]{seong2023leveraging}
Hyun~Seok Seong, WonJun Moon, SuBeen Lee, and Jae-Pil Heo.
\newblock Leveraging hidden positives for unsupervised semantic segmentation.
\newblock In \emph{Proceedings of the IEEE/CVF Conference on Computer Vision and Pattern Recognition}, 2023.

\bibitem[Shin et~al.(2022)Shin, Xie, and Albanie]{shin2022reco}
Gyungin Shin, Weidi Xie, and Samuel Albanie.
\newblock Reco: Retrieve and co-segment for zero-shot transfer.
\newblock \emph{Advances in Neural Information Processing Systems}, 35:\penalty0 33754--33767, 2022.

\bibitem[Tang et~al.(2020{\natexlab{a}})Tang, Huang, and Zhang]{tang2020long}
Kaihua Tang, Jianqiang Huang, and Hanwang Zhang.
\newblock Long-tailed classification by keeping the good and removing the bad momentum causal effect.
\newblock \emph{Advances in Neural Information Processing Systems}, 33:\penalty0 1513--1524, 2020{\natexlab{a}}.

\bibitem[Tang et~al.(2020{\natexlab{b}})Tang, Niu, Huang, Shi, and Zhang]{tang2020unbiased}
Kaihua Tang, Yulei Niu, Jianqiang Huang, Jiaxin Shi, and Hanwang Zhang.
\newblock Unbiased scene graph generation from biased training.
\newblock In \emph{Proceedings of the IEEE/CVF Conference on Computer Vision and Pattern Recognition}, pp.\  3716--3725, 2020{\natexlab{b}}.

\bibitem[Tang et~al.(2018)Tang, Perazzi, Djelouah, Ben~Ayed, Schroers, and Boykov]{tang2018regularized}
Meng Tang, Federico Perazzi, Abdelaziz Djelouah, Ismail Ben~Ayed, Christopher Schroers, and Yuri Boykov.
\newblock On regularized losses for weakly-supervised cnn segmentation.
\newblock In \emph{European Conference on Computer Vision}, pp.\  507--522, 2018.

\bibitem[Van Den~Oord et~al.(2017)Van Den~Oord, Vinyals, et~al.]{van2017neural}
Aaron Van Den~Oord, Oriol Vinyals, et~al.
\newblock Neural discrete representation learning.
\newblock \emph{Advances in Neural Information Processing Systems}, 30, 2017.

\bibitem[Van~Gansbeke et~al.(2021)Van~Gansbeke, Vandenhende, Georgoulis, and Van~Gool]{van2021unsupervised}
Wouter Van~Gansbeke, Simon Vandenhende, Stamatios Georgoulis, and Luc Van~Gool.
\newblock Unsupervised semantic segmentation by contrasting object mask proposals.
\newblock In \emph{Proceedings of the IEEE/CVF International Conference on Computer Vision}, pp.\  10052--10062, 2021.

\bibitem[Van~Gansbeke et~al.(2022)Van~Gansbeke, Vandenhende, and Van~Gool]{van2022discovering}
Wouter Van~Gansbeke, Simon Vandenhende, and Luc Van~Gool.
\newblock Discovering object masks with transformers for unsupervised semantic segmentation.
\newblock \emph{arXiv preprint arXiv:2206.06363}, 2022.

\bibitem[Vaswani et~al.(2017)Vaswani, Shazeer, Parmar, Uszkoreit, Jones, Gomez, Kaiser, and Polosukhin]{vaswani2017attention}
Ashish Vaswani, Noam Shazeer, Niki Parmar, Jakob Uszkoreit, Llion Jones, Aidan~N Gomez, {\L}ukasz Kaiser, and Illia Polosukhin.
\newblock Attention is all you need.
\newblock \emph{Advances in Neural Information Processing Systems}, 30, 2017.

\bibitem[Von~Luxburg(2007)]{von2007tutorial}
Ulrike Von~Luxburg.
\newblock A tutorial on spectral clustering.
\newblock \emph{Statistics and computing}, 17:\penalty0 395--416, 2007.

\bibitem[Wang et~al.(2022)Wang, Yi, Chen, and Zhu]{wang2022out}
Ruoyu Wang, Mingyang Yi, Zhitang Chen, and Shengyu Zhu.
\newblock Out-of-distribution generalization with causal invariant transformations.
\newblock In \emph{Proceedings of the IEEE/CVF Conference on Computer Vision and Pattern Recognition}, pp.\  375--385, 2022.

\bibitem[Wang et~al.(2020{\natexlab{a}})Wang, Huang, Zhang, and Sun]{wang2020visual}
Tan Wang, Jianqiang Huang, Hanwang Zhang, and Qianru Sun.
\newblock Visual commonsense r-cnn.
\newblock In \emph{Proceedings of the IEEE/CVF Conference on Computer Vision and Pattern Recognition}, pp.\  10760--10770, 2020{\natexlab{a}}.

\bibitem[Wang et~al.(2021{\natexlab{a}})Wang, Zhou, Yu, Dai, Konukoglu, and Van~Gool]{wang2021exploring}
Wenguan Wang, Tianfei Zhou, Fisher Yu, Jifeng Dai, Ender Konukoglu, and Luc Van~Gool.
\newblock Exploring cross-image pixel contrast for semantic segmentation.
\newblock In \emph{Proceedings of the IEEE/CVF International Conference on Computer Vision}, pp.\  7303--7313, 2021{\natexlab{a}}.

\bibitem[Wang et~al.(2021{\natexlab{b}})Wang, Zhang, Shen, Kong, and Li]{wang2021dense}
Xinlong Wang, Rufeng Zhang, Chunhua Shen, Tao Kong, and Lei Li.
\newblock Dense contrastive learning for self-supervised visual pre-training.
\newblock In \emph{Proceedings of the IEEE/CVF Conference on Computer Vision and Pattern Recognition}, pp.\  3024--3033, 2021{\natexlab{b}}.

\bibitem[Wang et~al.(2020{\natexlab{b}})Wang, Zhang, Kan, Shan, and Chen]{wang2020self}
Yude Wang, Jie Zhang, Meina Kan, Shiguang Shan, and Xilin Chen.
\newblock Self-supervised equivariant attention mechanism for weakly supervised semantic segmentation.
\newblock In \emph{Proceedings of the IEEE/CVF Conference on Computer Vision and Pattern Recognition}, pp.\  12275--12284, 2020{\natexlab{b}}.

\bibitem[Wen et~al.(2022)Wen, Zhao, Zheng, Zhang, and QI]{wen2022selfsupervised}
Xin Wen, Bingchen Zhao, Anlin Zheng, Xiangyu Zhang, and XIAOJUAN QI.
\newblock Self-supervised visual representation learning with semantic grouping.
\newblock In Alice~H. Oh, Alekh Agarwal, Danielle Belgrave, and Kyunghyun Cho (eds.), \emph{Advances in Neural Information Processing Systems}, 2022.

\bibitem[Xu et~al.(2015)Xu, Schwing, and Urtasun]{xu2015learning}
Jia Xu, Alexander~G Schwing, and Raquel Urtasun.
\newblock Learning to segment under various forms of weak supervision.
\newblock In \emph{Proceedings of the IEEE/CVF Conference on Computer Vision and Pattern Recognition}, pp.\  3781--3790, 2015.

\bibitem[Yang et~al.(2021{\natexlab{a}})Yang, Zhang, and Cai]{yang2021deconfounded}
Xu~Yang, Hanwang Zhang, and Jianfei Cai.
\newblock Deconfounded image captioning: A causal retrospect.
\newblock \emph{IEEE Transactions on Pattern Analysis and Machine Intelligence}, 2021{\natexlab{a}}.

\bibitem[Yang et~al.(2021{\natexlab{b}})Yang, Zhang, Qi, and Cai]{yang2021causal}
Xu~Yang, Hanwang Zhang, Guojun Qi, and Jianfei Cai.
\newblock Causal attention for vision-language tasks.
\newblock In \emph{Proceedings of the IEEE/CVF Conference on Computer Vision and Pattern Recognition}, pp.\  9847--9857, 2021{\natexlab{b}}.

\bibitem[Yin et~al.(2022)Yin, Wang, Wang, Xu, Zhang, Li, and Jin]{yin2022transfgu}
Zhaoyuan Yin, Pichao Wang, Fan Wang, Xianzhe Xu, Hanling Zhang, Hao Li, and Rong Jin.
\newblock Transfgu: a top-down approach to fine-grained unsupervised semantic segmentation.
\newblock In \emph{European Conference on Computer Vision}, pp.\  73--89. Springer, 2022.

\bibitem[Yue et~al.(2020)Yue, Zhang, Sun, and Hua]{yue2020interventional}
Zhongqi Yue, Hanwang Zhang, Qianru Sun, and Xian-Sheng Hua.
\newblock Interventional few-shot learning.
\newblock \emph{Advances in Neural Information Processing Systems}, 33:\penalty0 2734--2746, 2020.

\bibitem[Yue et~al.(2021)Yue, Wang, Sun, Hua, and Zhang]{yue2021counterfactual}
Zhongqi Yue, Tan Wang, Qianru Sun, Xian-Sheng Hua, and Hanwang Zhang.
\newblock Counterfactual zero-shot and open-set visual recognition.
\newblock In \emph{Proceedings of the IEEE/CVF Conference on Computer Vision and Pattern Recognition}, pp.\  15404--15414, 2021.

\bibitem[Zadaianchuk et~al.(2023)Zadaianchuk, Kleindessner, Zhu, Locatello, and Brox]{zadaianchuk2023unsupervised}
Andrii Zadaianchuk, Matthaeus Kleindessner, Yi~Zhu, Francesco Locatello, and Thomas Brox.
\newblock Unsupervised semantic segmentation with self-supervised object-centric representations.
\newblock In \emph{International Conference on Learning Representations}, 2023.

\bibitem[Zhang et~al.(2020{\natexlab{a}})Zhang, Zhang, Tang, Hua, and Sun]{zhang2020causal}
Dong Zhang, Hanwang Zhang, Jinhui Tang, Xian-Sheng Hua, and Qianru Sun.
\newblock Causal intervention for weakly-supervised semantic segmentation.
\newblock \emph{Advances in Neural Information Processing Systems}, 33:\penalty0 655--666, 2020{\natexlab{a}}.

\bibitem[Zhang et~al.(2021)Zhang, Torr, Ranftl, and Richter]{zhang2021looking}
Feihu Zhang, Philip Torr, Ren{\'e} Ranftl, and Stephan Richter.
\newblock Looking beyond single images for contrastive semantic segmentation learning.
\newblock \emph{Advances in Neural Information Processing Systems}, 34:\penalty0 3285--3297, 2021.

\bibitem[Zhang et~al.(2020{\natexlab{b}})Zhang, Jiang, Wang, Kuang, Zhao, Zhu, Yu, Yang, and Wu]{zhang2020devlbert}
Shengyu Zhang, Tan Jiang, Tan Wang, Kun Kuang, Zhou Zhao, Jianke Zhu, Jin Yu, Hongxia Yang, and Fei Wu.
\newblock Devlbert: Learning deconfounded visio-linguistic representations.
\newblock In \emph{ACM International Conference on Multimedia}, pp.\  4373--4382, 2020{\natexlab{b}}.

\bibitem[Zhou et~al.(2022)Zhou, Wei, Wang, Shen, Xie, Yuille, and Kong]{zhou2022image}
Jinghao Zhou, Chen Wei, Huiyu Wang, Wei Shen, Cihang Xie, Alan Yuille, and Tao Kong.
\newblock Image {BERT} pre-training with online tokenizer.
\newblock In \emph{International Conference on Learning Representations}, 2022.

\bibitem[Ziegler \& Asano(2022)Ziegler and Asano]{ziegler2022self}
Adrian Ziegler and Yuki~M Asano.
\newblock Self-supervised learning of object parts for semantic segmentation.
\newblock In \emph{Proceedings of the IEEE/CVF Conference on Computer Vision and Pattern Recognition}, pp.\  14502--14511, 2022.

\end{thebibliography}
\bibliographystyle{iclr2024_conference}

% Appendix
\appendix
%%%%%%%%%%%%%%%%%%%%%%%%%%%%%%%%%%%%%%%%%%%%%%%%%%%%%%%%%%%%%
% Appendix A
\section{Expansion of Related Works}
\label{appendix:A}

\paragraph{Unsupervised Semantic Segmentation.} 
One of the key challenges in unsupervised dense prediction is the need to learn semantic representations for each pixel without the guidance of labeled data. In an early work for unsupervised semantic segmentation (USS), \citet{ji2019invariant} introduced the IIC framework, which aims to maximize mutual information between feature representations obtained from augmented views. Subsequently, several methods have advanced the quality of segmentation by introducing inductive bias through cross-image correspondences~\citep{hwang2019segsort, cho2021picie, wen2022selfsupervised} or by incorporating saliency information in an end-to-end manner~\citep{van2021unsupervised, ke2022unsupervised}.

More recently, the discovery of semantic consistency in pre-trained self-supervised frameworks at the feature attention map~\citep{caron2021emerging} has led to prevalent approaches. \citet{hamilton2022unsupervised} introduced a method that leverages pre-trained knowledge and distills this information into the unsupervised segmentation task. Following this, various works~\citep{wen2022selfsupervised, yin2022transfgu, ziegler2022self} have employed self-supervised representations as pseudo segmentation labels~\citep{zadaianchuk2023unsupervised, liacseg} or as pre-encoded representations to incorporate additional prior knowledge~\citep{van2021unsupervised, zadaianchuk2023unsupervised} into the segmentation frameworks.

Our work aligns with~\citet{hamilton2022unsupervised, seong2023leveraging} in terms of enhancing segmentation features solely with the pre-trained representation. However, we emphasize the presence of indeterminate clustering targets inherent in unsupervised segmentation tasks. Our qualitative and quantitative results have demonstrated that the absence of evident clustering targets leads to poor segmentation outcomes in unsupervised settings. These negative effects have not been adequately addressed in previous works within the existing literature. Accordingly, for the first time, we interpret the unsupervised segmentation task within the context of causality, effectively bridging discretized representation learning and contrastive learning within this task.

\paragraph{Causal Inference.} In recent years, numerous studies~\citep{wang2020visual, zhang2020devlbert, scholkopf2021toward, lv2022causality} have applied causal inference techniques in DNNs to estimate the true causal effects between treatments and outcomes of interest. The fundamental approach to achieve causal identification involves blocking backdoor paths induced from confounders. 

In several computer vision methods have employed various causal approaches such as backdoor adjustment establishing explicit confounders~\citep{tang2020long,zhang2020causal,yue2020interventional,liu2022show}, mediation analysis~\citep{tang2020unbiased, niu2021counterfactual}, and generating counterfactual augmentations for randomized treatment assignments~\citep{agarwal2020towards, yue2021counterfactual,wang2022out} and have been successfully applied at task-specific levels. More recently, various works~\citep{kim2023demystifying, lee2023mitigating} have demonstrated that the causal approaches can be applied into the more specific computer vision areas with more sophisticated theories. 

However, one of the challenges of applying causal inference to computer vision tasks is the explicit definition of confounding variables and the full control of them. Accordingly, we utilize frontdoor adjustment~\citep{pearl1995causal} which can identify causal effects without the requirement of observed confounders, but relatively less explored in the context of computer vision tasks~\citep{yang2021causal, yang2021deconfounded}. Inspired by recent developments in discrete representation learning~\citep{van2017neural, esser2021taming}, we proactively build a discretized concept representation and serve it as an informative mediator, allowing us to establish criteria for identifying positive and negative samples for a given query pixel representation. Consequently, this approach facilitates the integration of discretized representation and self-supervised learning into the process of unsupervised semantic segmentation.
%%%%%%%%%%%%%%%%%%%%%%%%%%%%%%%%%%%%%%%%%%%%%%%%%

%%%%%%%%%%%%%%%%%%%%%%%%%%%%%%%%%%%%%%%%%%%%%%%%%%%%%%%%%%%%%
% Appendix B
\section{Detailed Implementaion of CAUSE}
\label{appendix:B}
We present a detailed description of a concrete implementation for CAUSE, expanding upon the algorithms outlined in the method section and providing additional implementation details not covered in the experiment section. To validate identifiable and reproducible performance described in the experiment section, one can access the trained parameters of CAUSE-MLP and CAUSE-TR, as well as their visual quality, through the code document available in the supplementary material.

\subsection{Maximizing Modularity}

When generating a mediator to design the concept cluster book, we need to compute a cluster assignment matrix $\mathcal{C}\in\mathbb{R}^{hw\times k}$ as described in Algorithm~\ref{alg:mediator} in our manuscript. However, a computational issue arises when $k$ becomes large, such as the value of $2048$ selected for the main experiments, thus computing the measure of \textit{Modularity} $\mathcal{H}$ becomes computationally expensive. To address this issue, we utilize the trace property of exchanging the inner multiplication terms, thereby reducing the computational burden, which can be written as follows:
\begin{equation}
    \mathcal{H} = \frac{1}{2e}\text{Tr}\left(\underbrace{\mathcal{C}(T, M)^\text{T}\left[\mathcal{A}-\frac{dd^\text{T}}{2e}\right]\mathcal{C}(T, M)}_{\mathbb{R}^{k\times k}}\right) = \frac{1}{2e}\text{Tr}\left(\underbrace{\mathcal{C}(T, M)\mathcal{C}(T, M)^\text{T}}_{\mathbb{R}^{hw \times hw}}\left[\mathcal{A}-\frac{dd^\text{T}}{2e}\right]\right).
\end{equation}
However, directly calculating the above formulation can lead to very small \textit{Modularity} due to multiplying very small two values from $\mathcal{C}(T, M)\mathcal{C}(T, M)^\text{T}$, rendering it ineffective optimization. To overcome this, we use the hyperbolic tangent and temperature term, which can be written as:
\begin{equation}
\mathcal{H} \approx \frac{1}{2e}\text{Tr}\left(\tanh\left(\frac{\mathcal{C}(T, M)\mathcal{C}(T, M)^\text{T}}{\tau}\right)\left[\mathcal{A}-\frac{dd^\text{T}}{2e}\right]\right),
\end{equation}
in order to scale up this value while it can maintain the possible range of the multiplication $\mathcal{C}(T, M)\mathcal{C}(T, M)^\text{T}$ from the following range:
\begin{equation}
    0\leq \tanh\left(\frac{\mathcal{C}(T, M)\mathcal{C}(T, M)^\text{T}}{\tau}\right)<1,
\end{equation}
such that it satisfies $0\leq \mathcal{C}(T, M)\mathcal{C}(T, M)^\text{T}\leq 1$ originated from the definition of the clustering assignment: $0\leq\mathcal{C}(T, M)=\max(0, \text{cos}(T, M))\leq 1$.

\subsection{Transformer-based Segmentation Head}

We use a single layer transformer decoder inspired by~\citet{vaswani2017attention, carion2020end} to build segmentation head with self-attention (SA), cross-attention (CA), and feed forward network (FFN) with its $2048$ inner-dimension by default hyper-parameter~\citep{vaswani2017attention}, where a single head attention is used on its enough performance. To explain the detailed propagation procedure for CAUSE-TR, we first show vector-quantization mechanism for the pre-trained feature representation $T=\{t\in\mathbb{R}^{c}\}^{hw}$ by replacing each patch feature point $t$ with each of the most closest concept features in $M$ as follows:
\begin{equation}
    Q=\{q\in\mathbb{R}^{c} \mid q=\arg\max_{m\in M}\text{cos}(t, m)\}^{hw}.
\end{equation}
Next, $Q$ is first propagated in SA, and $Q$ and $T$ are considered as query and key/value, respectively in CA, and the output of CA is propagated in FFN. Note that, learnable positional embedding is used in both query/key of SA and query of CA as \citet{carion2020end} have carried out. One different structure is to adopt additional two MLP layers in order to reduce the dimension from $c$ (ViT-S:384, ViT-B:768) to $r$ (90) for segmentation head output $Y$. This is because \citet{koenig2023uncovering} empirically demonstrate that higher dimension $r$ over $100$ brings in gradual collapse of clustering performance derived from the curse of dimensionality~\citep{assent2012clustering}.

\subsection{Anchor Selection}

In line 5 of Algorithm~\ref{alg:contrastive}, we describe that we sample anchor patch feature point $y$ in $Y=\{y\in\mathbb{R}^r\}^{hw}$. In reality, it is extremely inefficient to select all number $hw$ of the patch feature points in $Y$ to perform anchor points in self-supervised learning, because of the limitation of the resource-constrained hardware. Therefore, we use a high-computation-reduced technique of stochastic sampling only 6.25\% points ($\frac{1}{4}^2\times 100 (\%)$) among the number $hw$ points in $Y$, where we randomly select one feature point whenever a window having kernel size $4\times 4$ and stride $4$ is sliding along with $Y$.

\subsection{Positive \& Negative Concept Selection}
In line 6 of Algorithm~\ref{alg:contrastive}, we either describe that we sample positive and negative concept features $y^{+}, y^{-}$ in the set of $Y_{\text{ema}}$ and $Y_{\text{bank}}$ for the given anchor patch feature point $y$: it expresses $y^+, y^- \sim \{Y_{\text{ema}}, Y_{\text{bank}}\mid y\}$. First, we find the patch feature point $t$ corresponding to $y$ and then search the most closest concept $q=\arg\max_{m\in M}\text{cos}(t, m)$ and its index $\text{id}_q$. Next, we filter the positive $y^+$ and negative $y^-$ concept features satisfying each condition on $\mathcal{D}_M[\text{id}_q,:]>\phi^+$ and $\mathcal{D}_M[\text{id}_q,:]<\phi^-$. Then, we sample all of the positive and negative concept features in the set of $Y_{\text{ema}}$ and $Y_{\text{bank}}$. Note that, there are a few case that the row vector in $\mathcal{D}_M$ has a minimum value over zero, thus we technically set hard negative relaxation to $0.1$, instead of $0$.

\subsection{Concept Bank}

In line 10 of Algorithm~\ref{alg:contrastive}, the concept bank $Y_{\text{bank}}$ follows a specific rule: not all of the segmentation features $Y_{\text{ema}}$ are collected, but they are instead $50\%$ re-sampled based on the most closest concept indices individually, where the concept bank collects a maximum of $100$ features per concept prototype. Before re-sampling, $50\%$ of $Y_{\text{bank}}$ is randomly discarded. Considering that we have the number $2048$ of concept prototypes, the concept bank stores total number $100 \times 2048$ of the segmentation features. This ensures that the concept bank contains a comprehensive collection of treatment candidates $T'$, providing the diversity of selecting positive and negative concept features.

\subsection{Image Resolution and Augmentation}
For COCO-Stuff and Cityscapes, we equally follow data-augmentation method of STEGO~\citep{hamilton2022unsupervised} and HP~\citep{seong2023leveraging} which employ five-crop with crop ratio of $0.5$ in full image resolution and resizes the cropped images to $224\times 224$ for CAUSE-MLP in training phase. For inference phase, images are resized to $320\times 320$ along the minor axis followed by center crops of each validation image. For CAUSE-TR, $320\times 320$ image resolution is used to train segmentation head of a single layer transformer, because the same number of queries and learnable positional embeddings is used in training and inference phase. For Pascal VOC 2012, COCO-81, and COCO-171, we follow data-augmentation method of TransFGU~\citep{yin2022transfgu} which employs multiple-crop with multiple ratio. A significant different point is that STEGO, HP, and TransFGU employ additional data-augmentation techniques, including Horizontal Flip, Color-Jittering, Gray-scaling, and Gaussian-Blurring as geometric and photometric transforms, but CAUSE utilizes Horizontal Flip only.

%%%%%%%%%%%%%%%%%%%%%%%%%%%%%%%%%%%%%%%%%%%%%%%%%%%%%%%%%%%%%
% Appendix C
\section{Additional Experiments}
\label{appendix:C}
Due to page limitations, we are unable to include a comprehensive set of visual results for unsupervised semantic segmentation on multiple datasets in the main manuscript. In this additional section, we provide various examples primarily from four datasets and show the comparison results with baseline methods.

\subsection{Additional Qualitative Results}
To provide further evidence of unsupervised semantic segmentation results, we include additional qualitative visual results in Fig.\ref{fig:a_coco} and Fig.\ref{fig:a_city} for COCO-Stuff and Cityscapes, respectively. The entire experimental setup remains consistent with the main manuscript, and we compare our proposed method with recent unsupervised semantic segmentation baselines~\citep{hamilton2022unsupervised, seong2023leveraging, shin2022reco} that also utilize pre-trained DINO~\citep{caron2021emerging} feature representations.

Additionally, we present qualitative results for object-centric semantic segmentation by providing visualizations for the PASCAL VOC, COCO-81 and COCO-171 in Fig.~\ref{fig:pascal_coco} and Fig.~\ref{fig:coco171}, respectively. All of these datasets include an additional background class. While the negative relaxation is set to the same value of $0.1$, we have adjusted the positive relaxation to $0.2$, $0.4$, and $0.55$ for PASCAL VOC, COCO-81, and COCO-171 datasets, respectively. This modification is primarily due to account for the coarsely merged background class, as it aids in distinguishing the intricate integration of the background concepts.

\subsection{Failure Cases}
Unsupervised semantic segmentation is considered fundamentally challenging due to the combination of the absence of supervision and the need for dense pixel-level predictions. Even if we successfully achieve competitive unsupervised segmentation performance, there are some failure cases in which CAUSE may produce inadequate segmentation quality. 

\begin{figure*}[t]
\centering
\includegraphics[width=0.99\textwidth]{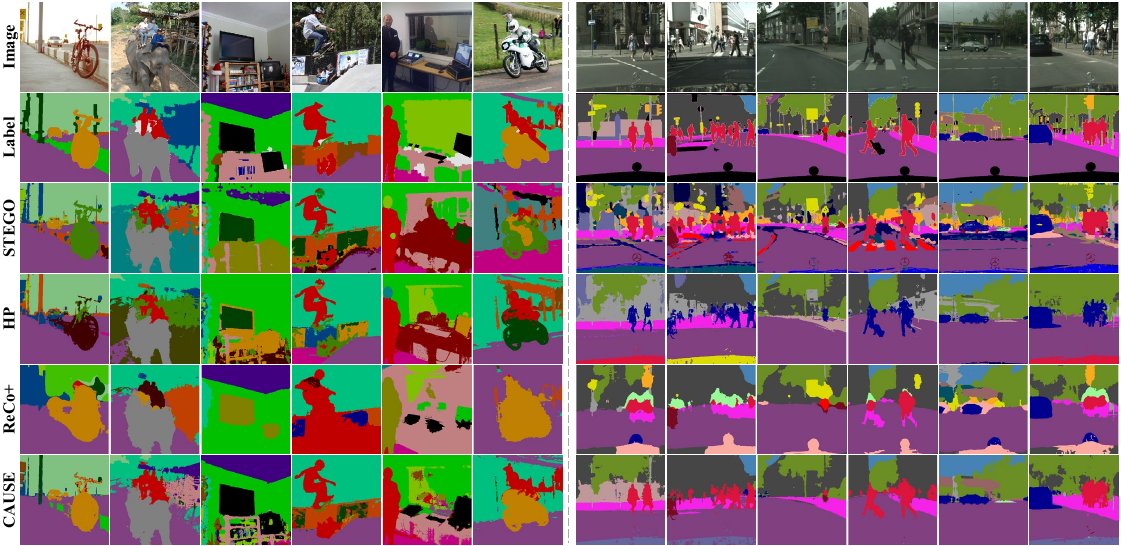}
\caption{Failure cases of CAUSE and comparison results with other baselines.} 
\label{fig:fail}

\end{figure*}

In Fig.~\ref{fig:fail}, we present failed segmentation with other baselines. One of the observations is the existence of noisy segmentation outcomes, especially in complex scenes. A straightforward solution is to adjust the larger number $k$ when constructing the concept clusterbook $M$. However, we have observed a trade-off between handling complex scenes and dealing with relatively easier examples. 
To fundamentally address this issue in future directions, we expect to explore pre-processing techniques and incorporate multi-scale feature extraction methods, as demonstrated in previous works such as~\citet{kirillov2019panoptic, kim2023universal, ranftl2021vision}. These approaches aim to enhance the precision of detailed dense prediction.

Additionally, we have observed instances where the cluster assignments were incorrectly predicted for certain object instances. We believe that these failures can primarily be attributed to two factors: (\lowercase\expandafter{\romannumeral1}) the inherent limitations of leveraging pre-trained self-supervised frameworks originally designed for image-level prediction tasks, and (\lowercase\expandafter{\romannumeral2}) the possibility of incompleteness in the employed clustering methods. As part of our future work, we believe it is essential to utilize foundation networks specifically designed for dense prediction tasks and clustering algorithms that can operate in an unsupervised manner. This approach will likely lead to more robust performance for unsupervised semantic segmentation.

\subsection{Concept Analysis in Clusterbook}
\begin{figure*}[t]
\centering
\includegraphics[width=0.9\textwidth]{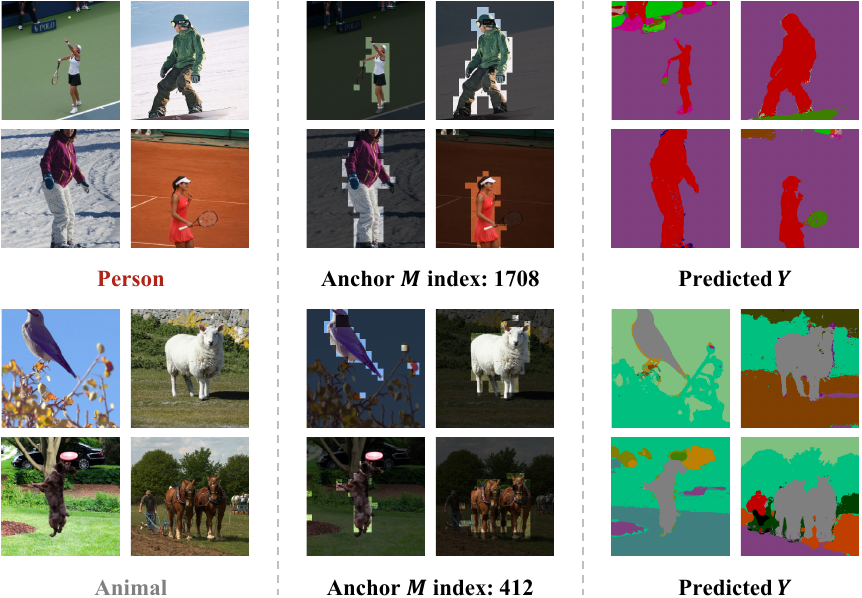}
\caption{Retrieval results of the concept with respect to the shared index on clusterBook. We select \textit{Person} and \textit{Animal} categories and CAUSE prediction results on COCO-Stuff.} 
\label{fig:concept}
\end{figure*}

In the proposed causal approach in unsupervised semantic segmentation, we define discretized representation as a mediator (\textit{concept clusterbook}) and leverage the advantages of discretization to facilitate concept-wise self-supervised learning through frontdoor adjustment. A natural question would be: \textit{What is included in the clusterbook within the representation space?} To address the question, we conduct additional experiments that focused on retrieving concepts using the shared index of clusterbook. Firstly, we select an anchor index from the total 2048 concepts in clusterbook. Then, we retrieve image regions that corresponds to the same cluster index as the anchor. Furthermore, to merge more wider image regions considering pixel-level clusterbook indices, we employ the concept distance matrix as explained in Section~\ref{sec:posneg}. Specifically, we merge image regions based on their discretized index when the concept distance with the anchor index exceeds positive relaxation of softly $0.3$. The retrieved results can be found in Fig.~\ref{fig:concept}.

% \paragraph{Unfair Evaluation.} To validate the performances of USS, PiCIE~\citep{cho2021picie}, TransFGU~\citep{yin2022transfgu}, STEGO~\citep{hamilton2022unsupervised}, HP~\citep{seong2023leveraging}, and ReCo+~\citep{shin2022reco} measure their mIoU and pAcc excepting `NaN' on category-wise performance, even though there are empirically lots of `NaN' across category-wise IoU and pAcc. Therefore, it induces unfair evaluation settings because once the number $\mathbb{C}-1$ of category-wise IoU is `NaN' and the last one category has $90\%$ IoU, then mIoU is just calculated to $90\%$ and this model becomes state-of-the-art model at once. To prevent this unfair evaluation, illustrating category-wise performance is an effective solution to identify genuine USS performances.

\section{Discussions and Limitations}
\label{appendix:D}
\begin{figure*}[b]
\centering
\begin{flushleft}
    \hspace{0.3cm} Img \hspace{0.4cm} GT \hspace{0.17cm} STEGO   \hspace{0.18cm} HP
    \hspace{0.23cm} ReCo+
    \hspace{0.07cm} \textbf{Ours}
    \hspace{0.5cm} Img \hspace{0.4cm} GT \hspace{0.17cm} STEGO   \hspace{0.18cm} HP
    \hspace{0.23cm} ReCo+
    \hspace{0.07cm} \textbf{Ours}
\end{flushleft}	
\vspace*{-0.2cm}
\includegraphics[width=0.99\textwidth]{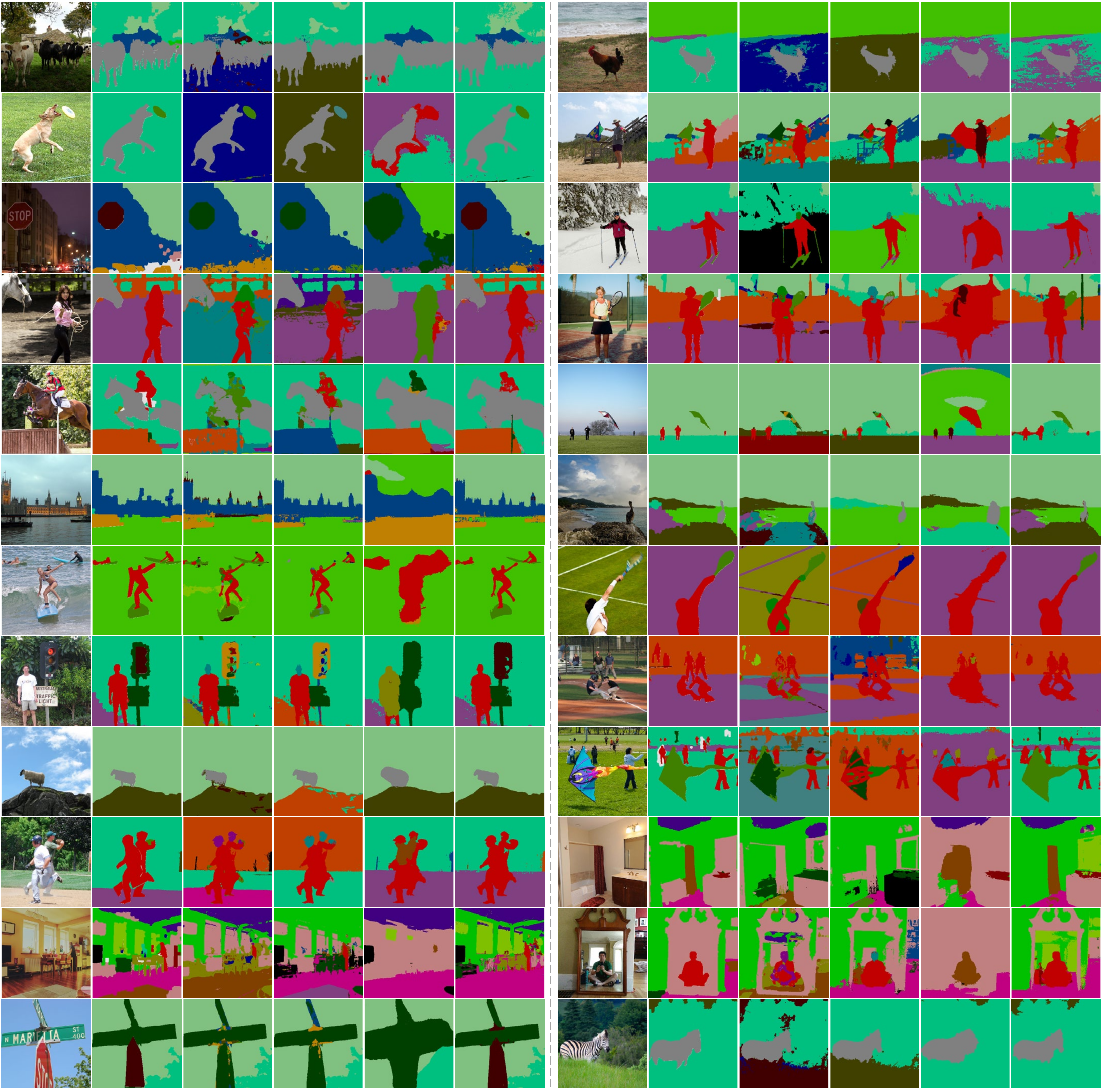}
\caption{Additional qualitative results of unsupervised semantic segmentation for Coco-Stuff. Please look up the object color maps in the main manuscripts.} 
\label{fig:a_coco}

\end{figure*}

\begin{figure*}[b]
\centering
\begin{flushleft}
    \hspace{0.3cm} Img \hspace{0.45cm} GT \hspace{0.15cm} CAUSE
    \hspace{0.1cm} Img \hspace{0.45cm} GT \hspace{0.15cm} CAUSE
    \hspace{0.27cm} Img \hspace{0.45cm} GT \hspace{0.15cm} CAUSE
    \hspace{0.1cm} Img \hspace{0.45cm} GT \hspace{0.15cm} CAUSE
\end{flushleft}		
\vspace*{-0.2cm}
\includegraphics[width=0.99\textwidth]{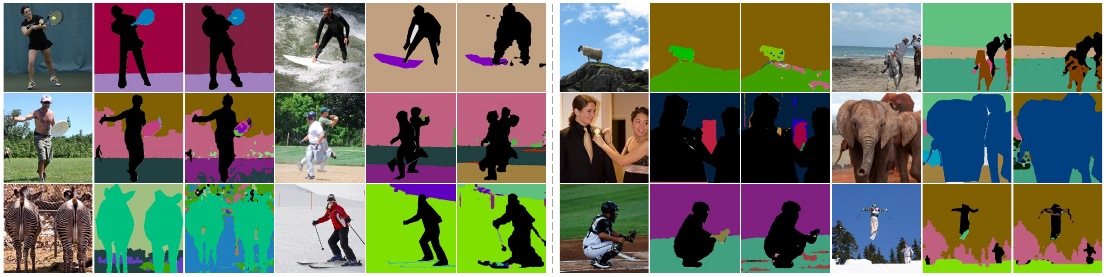}
\caption{Qualitative results of unsupervised semantic segmentation for COCO-171, which is specialized for object-centric semantic segmentation with 171 categories.} 
\label{fig:coco171}
\end{figure*}

\begin{figure*}[t]
\centering
\begin{flushleft}
    \hspace{0.3cm} Img \hspace{0.4cm} GT \hspace{0.17cm} STEGO   \hspace{0.18cm} HP
    \hspace{0.23cm} ReCo+
    \hspace{0.07cm} \textbf{Ours}
    \hspace{0.5cm} Img \hspace{0.4cm} GT \hspace{0.17cm} STEGO   \hspace{0.18cm} HP
    \hspace{0.23cm} ReCo+
    \hspace{0.07cm} \textbf{Ours}
\end{flushleft}		
\vspace*{-0.2cm}
\includegraphics[width=0.99\textwidth]{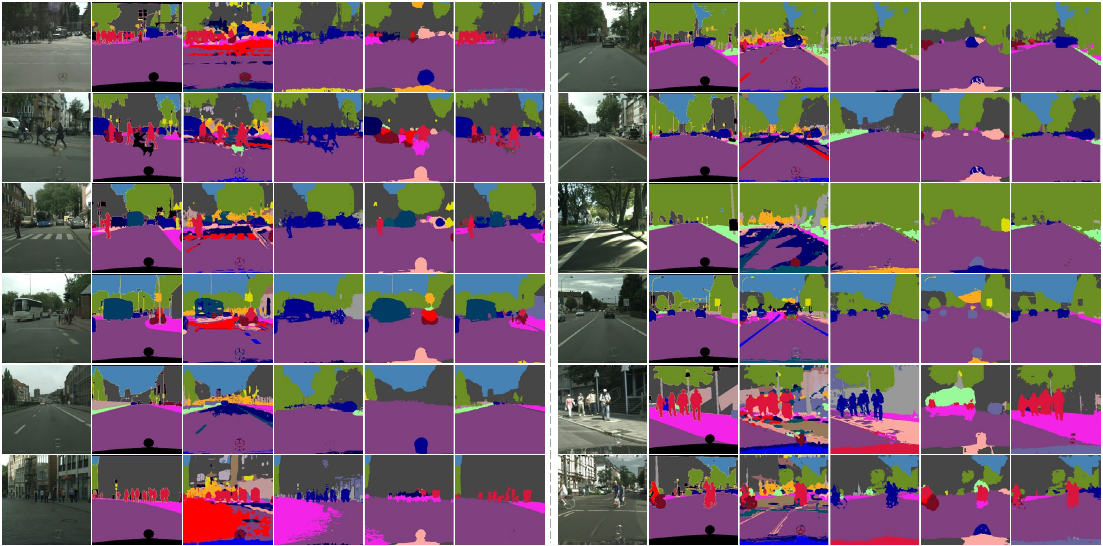}
\caption{Additional qualitative results of unsupervised semantic segmentation for Cityscapes. Please look up the object color maps in the main manuscripts.} 
\label{fig:a_city}

\end{figure*}

\begin{figure*}[t]
\centering
\begin{flushleft}
    \hspace{0.3cm} Img \hspace{0.45cm} GT \hspace{0.15cm} CAUSE
    \hspace{0.1cm} Img \hspace{0.45cm} GT \hspace{0.15cm} CAUSE
    \hspace{0.27cm} Img \hspace{0.45cm} GT \hspace{0.15cm} CAUSE
    \hspace{0.1cm} Img \hspace{0.45cm} GT \hspace{0.15cm} CAUSE
\end{flushleft}		
\vspace*{-0.2cm}
\includegraphics[width=0.99\textwidth]{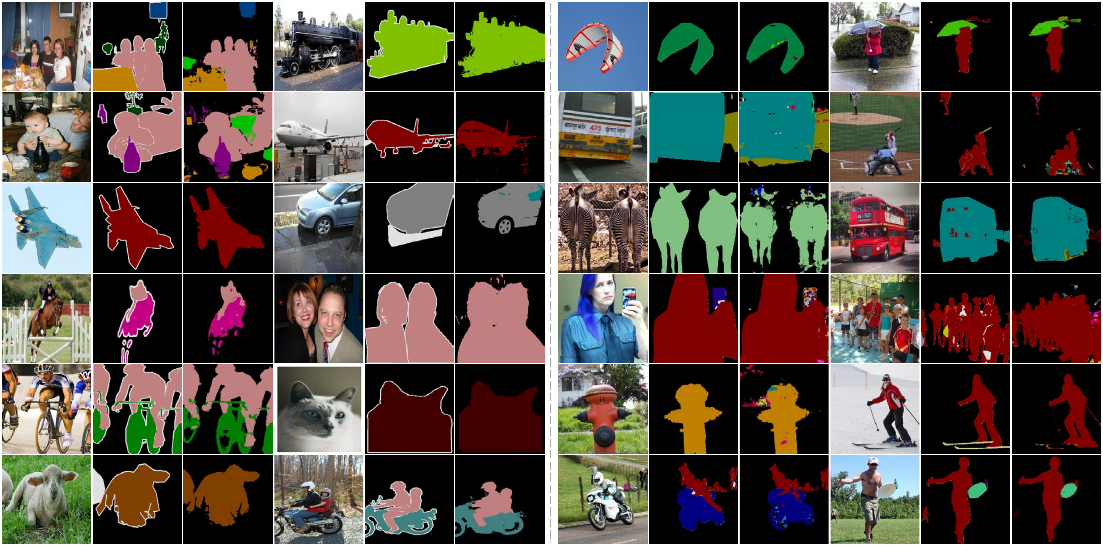}
\begin{flushleft}
    \hspace{2.1cm} (a) PASCAL VOC \hspace{4.5cm} (b) COCO-81
\end{flushleft}
\vspace*{-0.2cm}
\caption{Qualitative results of unsupervised semantic segmentation for PASCAL VOC and COCO-81, both of which are specialized for object-centric semantic segmentation.} 
\label{fig:pascal_coco}
\end{figure*}

\paragraph{Bootstrapping Pre-trained Models.} It is significantly challenging to handle fine-grained and complex scenes when dealing with unsupervised semantic segmentation using pre-trained feature representation. Based on the fact that the pre-trained features are designed to capture high-level semantic information, STEGO~\citep{hamilton2022unsupervised}, HP~\citep{seong2023leveraging}, and ReCo+~\citep{shin2022reco} struggle yet fail to precisely segment intricate details within images, especially in scenarios with densely packed objects, complex backgrounds, or small objects as observed in Fig.~\ref{fig:a_coco}-\ref{fig:pascal_coco}. 
This is because the pre-trained models, originally designed for tasks like image classification or object detection, are not perfectly matched to understand the different level of granularity required for fine-grained segmentation. In contrast, our novel framework, CAUSE, bootstraps the knowledge of high-level pre-trained feature representation to achieve semantic grouping at pixel-level by bridging a causal approach combining the discrete concept representation with concept-wise self-supervised learning.

\paragraph{Heuristically Static Hyper-Parameter.} 
CAUSE carefully assembles the concept clusterbook $M$, in advance, considering which concept features should be amalgamated or distinguished based on the intricate concept-wise distance matrix $\mathcal{D}_{M}$. One of nuisances involves heuristically establishing selection criterion for positive $\phi^+$ and negative $\phi^-$ relaxation, allowing for the construction of different level of granularity within the semantic groups $Y$. However, tailoring these criterion to the specifics of a dataset can be a challenging endeavor. Grid-search, ranging ambitiously from $0.1$ to $0.7$, is employed in the quest to find optimal relaxation values, but such task demands heuristic efforts. Moreover, in the realm of image processing, adapting to dynamic environmental contexts within images, encompassing scenarios such as the presence of small objects or intricate scenes, is imperative. In future direction, it requires more dynamical process of selecting criterion, particularly for such specialized and complex contexts.

% \newpage
%%%%%%%%%%%%%%%%%%%%%%%%%%%%%%%%%%%%%%%%%%%%%%%%%%%%%%%%%%%%%%%%%%%%%%%%%%%%%%%%%%%%
%%%%%%%%%%%%%%%%%%%%%%%%%%%%%%%%%%%%%%%%%%%%%%%%%%%%%%%%%%%%%%%%%%%%%%%%%%%%%%%%%%%%

%%%%%%%%%%%%%%%%%%%%%%%%%%%%%%%%%%%%%%%%%%%%%%%%%%%%%%%%%%%%%%%%%%%%%%%%%%%%%%%%%%%%
%%%%%%%%%%%%%%%%%%%%%%%%%%%%%%%%%%%%%%%%%%%%%%%%%%%%%%%%%%%%%%%%%%%%%%%%%%%%%%%%%%%%

%%%%%%%%%%%%%%%%%%%%%%%%%%%%%%%%%%%%%%%%%%%%%%%%%%%%%%%%%%%%%%%%%%%%%%%%%%%%%%%%%%%%
%%%%%%%%%%%%%%%%%%%%%%%%%%%%%%%%%%%%%%%%%%%%%%%%%%%%%%%%%%%%%%%%%%%%%%%%%%%%%%%%%%%%

%%%%%%%%%%%%%%%%%%%%%%%%%%%%%%%%%%%%%%%%%%%%%%%%%%%%%%%%%%%%%%%%%%%%%%%%%%%%%%%%%%%%
%%%%%%%%%%%%%%%%%%%%%%%%%%%%%%%%%%%%%%%%%%%%%%%%%%%%%%%%%%%%%%%%%%%%%%%%%%%%%%%%%%%%

%%%%%%%%%%%%%%%%%%%%%%%%%%%%%%%%%%%%%%%%%%%%%%%%%%%%%%%%%%%%%%%%%%%%%%%%%%%%%%%%%%%%
%%%%%%%%%%%%%%%%%%%%%%%%%%%%%%%%%%%%%%%%%%%%%%%%%%%%%%%%%%%%%%%%%%%%%%%%%%%%%%%%%%%%

%%%%%%%%%%%%%%%%%%%%%%%%%%%%%%%%%%%%%%%%%%%%%%%%%%%%%%%%%%%%%%%%%%%%%%%%%%%%%%%%%%%%
%%%%%%%%%%%%%%%%%%%%%%%%%%%%%%%%%%%%%%%%%%%%%%%%%%%%%%%%%%%%%%%%%%%%%%%%%%%%%%%%%%%%

\end{document}